\documentclass[10pt,twocolumn,letterpaper]{article}
\usepackage[table]{xcolor}
\usepackage{indentfirst}
\usepackage{amsfonts}
\usepackage{pifont}
\usepackage{circledsteps}
\usepackage{makecell}
\usepackage{multirow}
\usepackage{booktabs}
\usepackage{caption}
\usepackage{amsmath}
\usepackage{cancel}
\usepackage{bm}
\usepackage{circledsteps}
\usepackage[accsupp]{axessibility}

\usepackage[review]{cvpr}      

\definecolor{cvprblue}{rgb}{0.21,0.49,0.74}
\definecolor{hrefpink}{HTML}{ED06A1} 
\usepackage[pagebackref,breaklinks,colorlinks,allcolors=cvprblue]{hyperref}

\title{\textbf{A Self-Conditioned Representation Guided Diffusion Model for Realistic Text-to-LiDAR Scene Generation}}

\makeatletter
\renewcommand\thanks[1]{\footnotemark[\arabic{footnote}]\protected@xdef\@thanks{\@thanks
        \protect\footnotetext[\arabic{footnote}]{#1}}}
\makeatother

\author{Wentao Qu$^1$,
Guofeng Mei$^2$, 
Yang Wu$^1$,
YongShun Gong$^3$,
Xiaoshui Huang$^{4*}$,
Liang Xiao$^{1*}$ \vspace{3pt}
\\
NJUST$^1$, FBK‌$^2$, SDU$^3$, SJTU$^4$\\
\thanks{$^*$Corresponding Author. \; \href{https://github.com/QWTforGithub/CDSegNet}{\textcolor{hrefpink}{https://github.com/QWTforGithub/T2LDM}}}
}

\nolinenumbers

\date{}

\begin{document}
\maketitle




\begin{abstract}
Text-to-LiDAR generation can customize 3D data with rich structures and diverse scenes for downstream tasks. However, the scarcity of Text-LiDAR pairs often causes insufficient training priors, generating overly smooth 3D scenes. Moreover, low-quality text descriptions may degrade generation quality and controllability. In this paper, we propose a \textbf{T}ext-\textbf{to}-\textbf{L}iDAR \textbf{D}iffusion \textbf{M}odel for scene generation, named T2LDM, with a Self-Conditioned Representation Guidance (SCRG). Specifically, SCRG, by aligning to the real representations, provides the soft supervision with reconstruction details for the Denoising Network (DN) in training, while decoupled in inference. In this way, T2LDM can perceive rich geometric structures from data distribution, generating detailed objects in scenes. Meanwhile, we construct a content-composable Text-LiDAR benchmark, T2nuScenes, along with a controllability metric. Based on this, we analyze the effects of different text prompts for LiDAR generation quality and controllability, providing practical prompt paradigms and insights. Furthermore, a directional position prior is designed to mitigate street distortion,  further improving scene fidelity. Additionally, by learning a conditional encoder via frozen DN, T2LDM can support multiple conditional tasks, including Sparse-to-Dense, Dense-to-Sparse, and Semantic-to-LiDAR generation. Extensive experiments in unconditional and conditional generation demonstrate that T2LDM outperforms existing methods, achieving state-of-the-art scene generation.
\end{abstract}

\vspace{-24pt}
\section{Introduction}
\vspace{-3pt}

\begin{figure}[htp]
	\centering
	\includegraphics[width=0.48\textwidth]{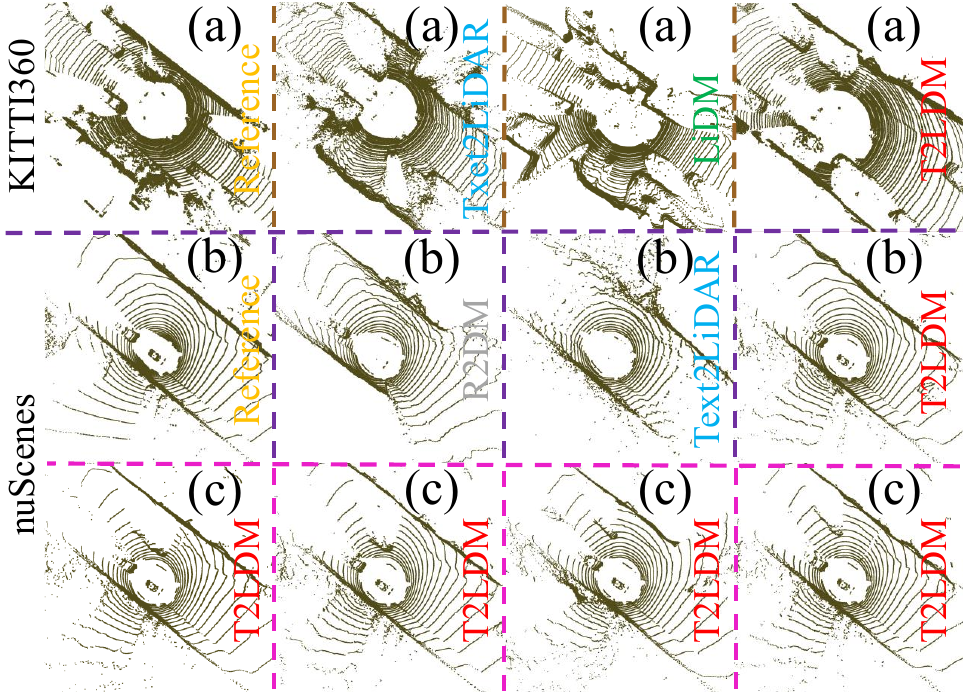}
	\vspace{-0.8cm}
	\caption{Due to lacking training priors, existing methods struggle to generate detailed scene objects. In contrast, T2LDM can generate realistic (a and b) and diverse (c) variants of the same scene.}
	\label{fig1}
	\vspace{-0.3cm}
\end{figure}



LiDAR perceives the surrounding environment and geometric structures, providing essential data support for 3D scene understanding tasks such as autonomous driving \cite{li2020deep}, virtual reality \cite{behari2025blurred}, and robotics \cite{yin2024survey}. However, collecting LiDAR scene data with diverse structures and under adverse weather conditions (e.g., rain) is costly \cite{bijelic2020seeing, nakashima2024lidar, wu2024text2lidar}, limiting the advance of data-driven 3D perception models. Therefore, this has motivated growing research interest in synthesizing realistic, diverse and controllable LiDAR scenes.


Convenient natural language can provide semantic guidance for controllable scene generation. In recent years, numerous studies have achieved success in Text-to-Image generation tasks. Benefiting from rich Text-Image paired data \cite{schuhmann2022laion, coyo700m2023}, some methods can be trained on even over 100 million samples \cite{ramesh2022hierarchical, saharia2022photorealistic, rombach2022high}. This provides strong cross-modal alignment and semantic priors, enabling generative models to synthesize realistic and diverse visual content from natural language descriptions. Inspired by this, researchers attempt to introduce text conditions for customizing LiDAR scenes \cite{wu2024text2lidar}, improving 3D data diversity and scalability.

Unfortunately, unlike the easily collected Text-Image pairs (\emph{e.g.}, the internet \cite{radford2021learning, coyo700m2023}), LiDAR data acquisition is time-consuming and labor-intensive, making high-quality and diverse Text-LiDAR pairs extremely scarce (the Text-LiDAR pairs $<$ 35K in nuScenes \cite{caesar2020nuscenes}). This limitation hinders the sufficient training of generative models, thus often resulting in overly smooth and homogeneous generation results that lack distinct object structures and realistic details in LiDAR scenes (see Fig.~\ref{fig1}). Moreover, the content and form of text prompts for describing scene structures are crucial to generation results \cite{liu2022design, oppenlaender2025prompting}. However, existing benchmarks \cite{caesar2020nuscenes} only provide unnatural text descriptions and lack controllability evaluation metrics, further limiting the generation results of Text-to-LiDAR generative models.

To address these problems, in this paper, we propose a Text-to-LiDAR Diffusion Model , named T2LDM. Inspired by injecting regularization into DDPMs through representation learning \cite{li2023self, yu2024representation}, T2LDM employs a Self-Conditional Representation Guidance (SCRG) to learn geometric details from data distribution, improving object fidelity in generated LiDAR scenes. Specifically, SCRG leverages a Guidance Network (GN) to perceive multi-scale perturbed features from DN while aligning with the real representations. This allows GN to produce geometrically detailed features under multi-level perturbations and conditional guidance, providing multi-scale supervision signals for DN. In this way, T2LDM can effectively learn geometric structures from data distribution, generating detailed and realistic objects in scenes (see Fig.~\ref{fig1}). Moreover, unlike requiring pretrained priors \cite{li2023self, yu2024representation}, SCRG operates in an end-to-end paradigm, and \textit{\textbf{GN participates in gradient backpropagation only during the early training stage  while detached during inference,}} alleviating the computational cost.

Meanwhile, we construct a content-composable Text-LiDAR benchmark, T2nuScenes, with 3D box priors. This offers three advantages. First, this enables more accurate description of object locations than manual annotations (accuracy). Second, this can generalize to any 3D detection dataset (generality). Third, this enables controllability evaluation via detectors (evaluability). Moreover, \textit{\textbf{Text-to-LiDAR generation presents more flexibility than conditioning on complex 3D boxes.}} This is also the first to explore 3D box priors for scene text description, encouraging more explorations of 3D text-guided scene generation. Based on this, we  investigate the effect of text forms for LiDAR generation, providing prompt paradigms and insights.


Furthermore, we find that the spherical projection from LiDAR data to range map may cause \textit{\textbf{directional confusion}}, leading to distorted streets in generated scenes (see Fig.~\ref{fig2}(c)). Therefore, we design a directional position encoding to provide T2LDM with true directional priors for rows and columns of range maps, further improving fidelity.

Additionally, T2LDM exhibits excellent results across various conditional generation tasks via \textit{\textbf{non-latent ControlNet \cite{zhang2023adding}}}. Our key contributions can be summarized as:
\begin{itemize}
	\item We propose a Text-to-LiDAR Diffusion Model,  T2LDM, with a self-conditioned representation guidance.
	\item We construct a high-quality content-composable Text-LiDAR benchmark, T2nuScenes, exploring effective text prompt forms and providing insights.
	\item By leveraging a directional position prior, T2LDM alleviates road distortion, further improving scene fidelity.
	\item Unconditional and conditional results show that T2LDM can generate LiDAR scenes with detailed objects.
	
\end{itemize}

\section{Related Works}
\noindent\textbf{LiDAR Scene Generation.} LiDAR data can precisely describe the geometric structures and spatial relationships for a scene, providing an effective representation of the real-world environment. However, the high acquisition cost, labor-intensive annotation process, and rarity of adverse weather conditions make high-quality and diverse LiDAR data extremely scarce \cite{caccia2019deep, zyrianov2022learning, nakashima2024lidar, wu2024text2lidar}. Some methods attempt to synthesize realistic scenes on LiDAR data using physics-based simulation \cite{manivasagam2020lidarsim, hahner2021fog, teufel2022simulating, yang2024realistic}. They typically model the physics of LiDAR signals based on optical scattering and laser propagation principles, simulating signal attenuation, backscattering, and measurement noise. Although effective, directly simulating realistic scenes on LiDAR data struggles to generate geometrically diverse and structurally rich scenes, as this requires high-quality LiDAR data as the shape foundation. Benefiting from the strong data-driven capability of deep learning, some researchers have explored using neural networks to generate structurally diverse LiDAR scenes. \cite{caccia2019deep} is the first to explore LiDAR scene generation using generative models (VAE \cite{kingma2013auto} and GAN \cite{goodfellow2014generative}), exhibiting promising results. Inspired by this, some works use DDPMs \cite{ho2020denoising} to generate 3D scenes \cite{nakashima2024lidar, ran2024towards, wu2024text2lidar}, achieving superior performance. 

Although existing methods have achieved promising LiDAR scene generation results, insufficient training priors often lead to overly smooth and homogeneous scenes, limiting applicability. In this paper, we propose a Self-Conditional Representation Guidance, encouraging DN to learn geometric representations by regularization, improving object details and structural fidelity in generated scenes.


\begin{table*}[h]
	\resizebox{1.0\textwidth}{!}{
			\begin{tabular}{p{1.5cm}|p{2.0cm}|p{8.3cm}p{4.0cm}|p{1.4cm}p{1.4cm}p{1.5cm}}	
				\Xhline{1pt}
				
				{Level}
				&{Type}
				&{Prompt Example}
				&{Sample Distribution}
				&\makecell[c]{FSID $\downarrow$}
				&\makecell[c]{FPVD $\downarrow$}
				&\makecell[c]{TBR($\%$) $\uparrow$}\\
				
				\Xhline{1pt}
				
				\multirow{10}{*}{($\color[rgb]{1.0,0.9020,0}{\blacksquare}$) Object}
				&\multirow{5}{*}{Quantity}
				&{($\color{blue}{\spadesuit}$) Two cars.}
				&{857,1873,2099,29320}
				&\makecell[c]{67.32}
				&\makecell[c]{66.34}
				&\makecell[c]{31.12}\\
				
				&
				&{($\color{blue}{\spadesuit}$) There are two cars in the scene.}
				&{857,1873,2099,29320}
				&\makecell[c]{67.54}
				&\makecell[c]{66.78}
				&\makecell[c]{30.55}\\
				
				&
				&{($\color{blue}{\spadesuit}$) Two cars. One car is in front. One car is behind.}
				&{857,1873,2099,29320}
				&\makecell[c]{67.63}
				&\makecell[c]{66.81}
				&\makecell[c]{29.88}\\
				
				&
				&{($\color{red}{\clubsuit}$) Two cars. $\rightarrow$ There are two cars in the scene.}
				&{857,1873,2099,29320}
				&\makecell[c]{68.55}
				&\makecell[c]{67.10}
				&\makecell[c]{29.23}\\
				
				&
				&{($\color{red}{\clubsuit}$) There are two cars in the scene. $\rightarrow$ Two cars.}
				&{857,1873,2099,29320}
				&\makecell[c]{68.32}
				&\makecell[c]{67.01}
				&\makecell[c]{29.94}\\
				
				&\cellcolor[rgb]{0.8549, 0.9098, 0.9882} Adjusted Text
				&\cellcolor[rgb]{0.8549, 0.9098, 0.9882} {($\color[rgb]{0.4392,0.1882,0.6275}{\blacklozenge}$) Less/More than five cars.}
				&\cellcolor[rgb]{0.8549, 0.9098, 0.9882} {10692,23457}
				&\cellcolor[rgb]{0.8549, 0.9098, 0.9882} \makecell[c]{65.10}
				&\cellcolor[rgb]{0.8549, 0.9098, 0.9882} \makecell[c]{64.15}
				&\cellcolor[rgb]{0.8549, 0.9098, 0.9882} \makecell[c]{60.35}\\
				\cline{2-7}
				
				&{Location}
				&{One car is behind to the right of one pedestrian.}
				&{681,478,...,339,468}
				&\makecell[c]{68.12}
				&\makecell[c]{66.95}
				&\makecell[c]{12.23}\\
				
				&\cellcolor[rgb]{1.0, 0.9020, 0.8} Adjusted Text
				&\cellcolor[rgb]{1.0, 0.9020, 0.8} {($\color[rgb]{0.4392,0.1882,0.6275}{\blacklozenge}$) No car./One car is around one pedestrian/barrier/truck.}
				&\cellcolor[rgb]{1.0, 0.9020, 0.8} {12273,11534,3819,6523}
				&\cellcolor[rgb]{1.0, 0.9020, 0.8} \makecell[c]{66.74}
				&\cellcolor[rgb]{1.0, 0.9020, 0.8} \makecell[c]{65.54}
				&\cellcolor[rgb]{1.0, 0.9020, 0.8} \makecell[c]{23.42}\\
				
				\cline{2-7}
				
				&{Orientation}
				&{One car is facing backward.}
				&{7416,6662,9994,8711,1366}
				&\makecell[c]{66.45}
				&\makecell[c]{65.12}
				&\makecell[c]{37.12}\\
				
				&\cellcolor[rgb]{0.8352, 0.9098, 0.8314} Adjusted Text
				&\cellcolor[rgb]{0.8352, 0.9098, 0.8314} {($\color[rgb]{0.4392,0.1882,0.6275}{\blacklozenge}$) No car./One car is facing right/left.}
				&\cellcolor[rgb]{0.8352, 0.9098, 0.8314} {1366,16127,16656}
				&\cellcolor[rgb]{0.8352, 0.9098, 0.8314} \makecell[c]{65.54}
				&\cellcolor[rgb]{0.8352, 0.9098, 0.8314} \makecell[c]{64.32}
				&\cellcolor[rgb]{0.8352, 0.9098, 0.8314} \makecell[c]{59.42}\\
				
				\hline
				
				\multirow{2}{*}{($\color[rgb]{0.9725,0.3216,0.8627}{\blacktriangle}$) Scene}
				&{Weather}
				&{Rainy.}
				&{6670,27479}
				&\makecell[c]{65.14}
				&\makecell[c]{64.55}
				&\makecell[c]{-}\\
				
				\cline{2-7}
				
				&{Time}
				&{Night.}
				&{3987,30162}
				&\makecell[c]{65.53}
				&\makecell[c]{64.74}
				&\makecell[c]{-}\\
				
				\hline
				
				{-}
				&\cellcolor[rgb]{1.0, 0.9490, 0.8} {Wea., Loc.}
				&\cellcolor[rgb]{1.0, 0.9490, 0.8} {($\color[rgb]{0.4392,0.6784,0.2784}{\bigstar}$) Rainy. One car is around one pedestrian}
				&\cellcolor[rgb]{1.0, 0.9490, 0.8} {10101,9876,2966,4536,...}
				&\cellcolor[rgb]{1.0, 0.9490, 0.8} \makecell[c]{66.93}
				&\cellcolor[rgb]{1.0, 0.9490, 0.8} \makecell[c]{65.84}
				&\cellcolor[rgb]{1.0, 0.9490, 0.8} \makecell[c]{23.44}\\
				
				\hline
				
				\Xhline{1pt}
				
			\end{tabular}
		}
		\vspace{-4pt} 
		\caption{Results of different text forms. "Text1. $\rightarrow$ Text2." means that the model is trained with the text form of "Text2", while using "Text1" as conditional input in inference. "Wea., Loc." denotes "Weather, Location". In nuScenes, original descriptions like “\textit{\textbf{Turn right at intersection, cross bridge, many peds}}” are unnatural (the comparison between the original and re-annotated text descriptions in \textcolor{blue}{SM}).} 
		\label{tab31}
		\vspace{-3mm}
	\end{table*}

\noindent\textbf{Text-Guided Generation.} Natural language can intuitively describe scene content, providing flexible semantic guidance. Benefiting from the available and high-quality Text-Image sample pairs \cite{schuhmann2022laion, coyo700m2023}, many methods can successfully generate semantically aligned and diverse high-fidelity images from given natural language descriptions \cite{ramesh2022hierarchical, saharia2022photorealistic, rombach2022high}. Inspired by these advances, some researchers have explored using text guidance to generate 3D data. Due to the lack of high-quality Text-Point Cloud data, early methods leverage Text-Image priors to bridge the gap between text and point clouds, achieving object-level Text-to-Point Cloud generation \cite{nichol2022point, poole2022dreamfusion, lin2023magic3d}. Subsequently, several works attempted to directly generate object-level 3D data from text prompts \cite{luo2021diffusion, wu2023sketch}. Recently, some methods have showed the promise of Text-to-LiDAR scene generation \cite{wu2024text2lidar}.
	
Some explorations have demonstrated the potential of text-guided LiDAR generation, but the lack of high-quality Text-LiDAR pairs hinder this progress. In this paper, we construct a content-composable Text-LiDAR benchmark and provide a controllability evaluation metric. This relies only on 3D box priors, enabling easy extension to detection datasets and promoting Text-to-LiDAR generation research.

\section{Text Prompt for LiDAR Generation}
In this section, we use the re-annotated Text-LiDAR benchmark to evaluate the effect of different text prompts for LiDAR generation quality and controllability, providing the optimized prompt form and scene description insights.

\vspace{-1pt}
\subsection{Annotation and evaluation}
\vspace{-2pt}

\label{sec31}

\noindent\textbf{Text Annotation.} We re-annotated all LiDAR data from 34149 samples in nuScenes with object-level (the target object: “car”) and scene-level text descriptions (the annotation process in the supplementary material (\textcolor{blue}{SM})). Meanwhile, they are stored independently for flexible text combinations.

\noindent\textbf{Evaluation Metric.} We train T2LDM on different text forms, using FID \cite{ran2024towards} and TBR to evaluate generation quality and controllability. TBR means the matching \textbf{R}ate between the \textbf{T}ext prompt and the \textbf{B}oxes obtained by applying a detector \cite{liu2025fshnet} on 10,000 generated samples (details in \textcolor{blue}{SM}).

\vspace{-1pt}
\subsection{Text Prompt Comparison}
\vspace{-2pt}

\noindent\textbf{Quantity, Location, and Orientation ($\color[rgb]{1.0,0.9020,0}{\blacksquare}$).} Tab.~\ref{tab31} shows the comparison of generation quality and controllability across different object-level text prompts. Surprisingly, the explicit location prompt yields the worst results. Moreover, the generation controllability is substantially lower than that of other text prompt forms.

\noindent\textbf{Weather and Time ($\color[rgb]{0.9725,0.3216,0.8627}{\blacktriangle}$).} Meanwhile, we also exhibit the results of scene-level text descriptions in Tab.~\ref{tab31}, significantly outperforming object-level text prompts in generation.  

\noindent\textbf{Text Length ($\color{blue}{\spadesuit}$).} We further evaluate the generation results for different text lengths. In Tab.~\ref{tab31}, longer text prompts with similar semantics cause a slight degradation in results. This is because, \textit{\textbf{redundant information in longer prompts may hinder the model from capturing key semantics \cite{liu2022design}.}}

\noindent\textbf{Form Transfer ($\color{red}{\clubsuit}$).} Furthermore, as shown in Tab.~\ref{tab31}, text prompts with similar semantics but different forms cause only a slight decrease in results. We believe that \textit{\textbf{this benefits from the text encoder \cite{radford2021learning} effectively identifying semantics and producing reliable features.}}

\subsection{Text Prompt Analysis} 
\vspace{-2pt}
\label{sec33}


In general, layout-aware text prompts are intuitively expected to enhance generation quality and controllability \cite{zhou2024layout, zhou2025layoutdreamer}. However, the results in Tab.~\ref{tab31} indicate otherwise. In fact, this phenomenon can be explained from the perspective of \textit{\textbf{sample distribution}}. The more dispersed sample distribution can produce richer text descriptions but may cause insufficient training priors due to sample scarcity. This becomes particularly severe when the training data are inadequate. This also provides an explanation: \textit{\textbf{the more complex texts typically lead to poorer generation results \cite{liu2022compositional, podell2023sdxl}}}. 

Meanwhile, the above results provide some insights:
\begin{itemize}
	\item Text prompts should be clear and concise while retaining sufficient semantic information (see \textbf{Text Length}).
	\item A strong semantic-aware text encoder is crucial for text prompt generalization \cite{saharia2022photorealistic} (see \textbf{Form Transfer}).
	\item Annotating Text-LiDAR sample pairs should account for the sample distribution of the dataset to generate appropriate text descriptions for each scene (see Sec.~\ref{sec33}).
	
\end{itemize}

\vspace{-3pt}
\subsection{Text Prompt Optimization}
\vspace{-4pt}

\label{sec34}

\noindent\textbf{Adjusting Text ($\color[rgb]{0.4392,0.1882,0.6275}{\blacklozenge}$).} Based on the above insights, we adjust the description forms of quantity, position, and orientation in Tab.~\ref{tab31}. The improvement in generation quality and controllability shows the reliability of annotating text descriptions from the sample distribution perspective.

\noindent\textbf{Prompt Template ($\color[rgb]{0.4392,0.6784,0.2784}{\bigstar}$).} Based on the sample distribution and text diversity, we consider the benchmark prompt as “weather, location”, since this covers \textit{\textbf{the weather, object number, and object layout}} of scenes. We also provide detailed sample distributions of text combinations in \textcolor{blue}{SM}.

\vspace{-5pt}
\section{Methodology}

\subsection{Generation Process}
\label{sec41}

\noindent\textbf{Input Representation.} Range Map (RM) represents the entire LiDAR scene through a spherical projection of 3D coordinates \cite{milioto2019rangenet++}. The columns and rows represent the LiDAR \textbf{H}orizontal  ($0^\circ$-$360^\circ$, HFoV) and \textbf{V}ertical ($f_{down}$-$f_{up}$, VFoV) \textbf{F}ields \textbf{o}f \textbf{V}iew. The projection of $\bm{p}_i=(x,y,z)$ is:

\vspace{-10pt}
\begin{equation}
	\begin{split}
		\label{f411}
		u = \frac{1}{2}[1-arctan(y,x)\pi^{-1}]W, \quad\;\;\; \\
		v = [1-(arcsin(zr^{-1})+f_{up})f^{-1}]H, 
	\end{split}
\end{equation}
where $(u, v)$ and $(H, W)$ denote the 2D coordinates and the height, width of RM. Meanwhile, $r=||\bm{p}_i||^2$ represents the depth distance of each point $\bm{p}_i$ from the LiDAR sensor.

Range Map and LiDAR data exhibit a (partially) invertible relationship (LiDAR $\rightarrow$ RM $\rightarrow$ LiDAR), thus can be used for generation tasks. Meanwhile, we use depth $r$ and intensity $I$ as pixel values of RM$\in \mathbb{R}^{H \times W \times 2}$ \cite{nakashima2024lidar, wu2024text2lidar}.

\noindent\textbf{Conditional DDPMs for Text-to-LiDAR Generation.} Given a Range Map $\bm{x_0} \sim \mathcal{P}_{RM}$ projected from LiDAR coordinates by Eq.~\ref{f411}, a conditional text $\bm{c} \sim \mathcal{P}_{text}$, and a prior noise $\bm{x_T} \sim \mathcal{P}_{noise}$, conditional DDPMs achieve the distribution transformation process between $\mathcal{P}_{RM}$ and $\mathcal{P}_{noise}$ via: a predefined forward process $q$ that gradually adds perturbation to $\bm{x_0}$ until $\bm{x_T}$, and a trainable reverse process $p_\theta$ that slowly removes noise $\bm{x_T}$ back to $\bm{x'_0}$ conditioned on $\bm{c}$. Meanwhile, the timestep $t \sim \mathcal{U}[1024]$ governs the transition dynamics. In this process, \textit{\textbf{to effectively learn the distribution transformation, conditional DDPMs typically require sufficient training priors to match $\mathcal{P}_{RM}$ \cite{wang2023patch, zhu2025domainstudio}.}} 


Then, the training objective of conditional DDPMs is:

\vspace{-10pt}
\begin{equation}
	\begin{split}
		\label{f412}
		L(\theta) =
		\mathbb{E}_{\bm{\epsilon} \sim \mathcal{N}(0,I)}||\bm{v} - v_\theta(\bm{x_t},t,\bm{c})||^2, 
	\end{split}
\end{equation}
where the target $\bm{v}$ can be converted into $\bm{\epsilon}$ or $\bm{x_0}$ (the derivation in \textcolor{blue}{SM}). Meanwhile, unconditional generation can be regarded as a special case of conditional generation ($\bm{c}=\emptyset$) \cite{qu2024conditional}. Therefore, in this paper, we can achieve the classifier-free guidance (CFG) \cite{ho2022classifier} by alternately training unconditional DDPMs and conditional DDPMs. 

Subsequently, we can iteratively transform $\bm{x_T}$ sampled from $\mathcal{P}_{noise}$ to $\bm{x'_0} \sim \mathcal{P}_{RM}$ by the trained $v_\theta$ in inference. 

Finally, $\bm{x'_0}$ is converted back to the 3D coordinates to generate the LiDAR scene using the inverse of Eq.~\ref{f411} \cite{milioto2019rangenet++}.

\begin{figure}[htp]
	\centering
	\includegraphics[width=0.48\textwidth]{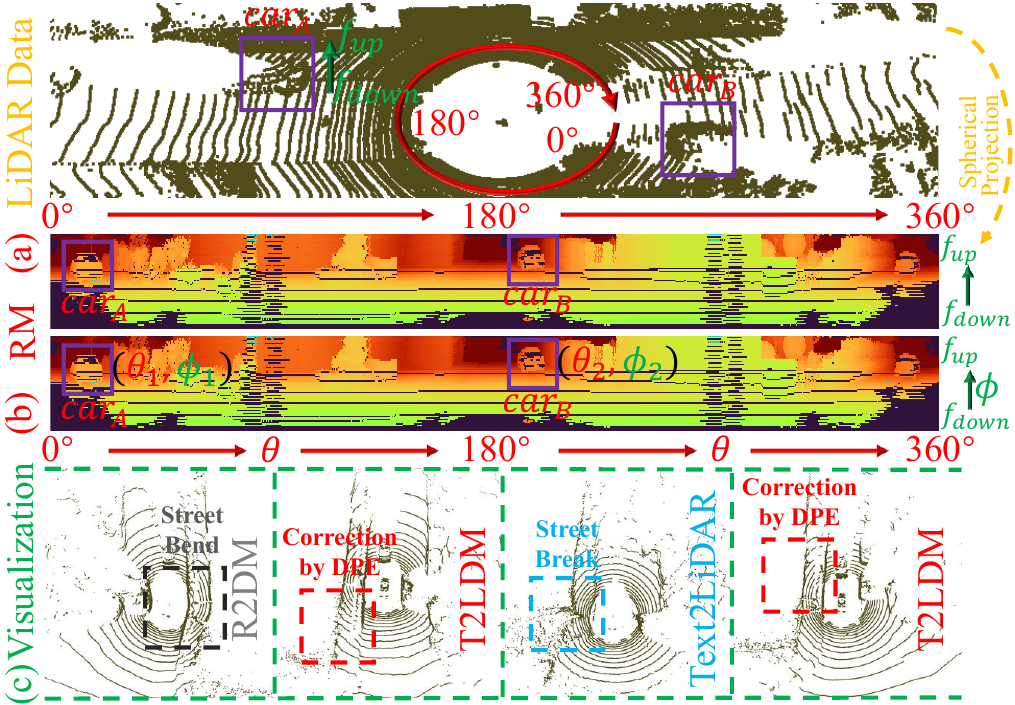}
	\vspace{-0.7cm}
	\caption{(a) In LiDAR space, $car_A$ is at \textit{\textbf{the front-right}} of $car_B$, but it appears as \textit{\textbf{the left}} in RM (window shift). (b) By defining the horizontal angle $\theta$ and vertical angle $\phi$ in RM, DPE provides true directional priors, enabling the model to correctly perceive object orientations in the scene. For example, the model can clearly understand the relative position between $car_A$($\theta_1$,$\phi_1$) and $car_B$($\theta_2$,$\phi_2$). (c) Existing methods produce bend or broken streets due to \textit{\textbf{directional confusion}}. T2LDM generates realistic ones.}
	\label{fig2}
	
\end{figure}

\subsection{Self-Conditioned Representation Guidance}

\label{sec42}

Unlike easily accessible Text-Image data \cite{li2023self, yu2024representation}, Text-LiDAR pairs are scarce due to costly collection and annotation \cite{caccia2019deep, zyrianov2022learning, nakashima2024lidar, wu2024text2lidar}. This often leads to insufficient training priors for generative models, resulting in overly smooth results that lack detailed objects in LiDAR scenes (see Fig.~\ref{fig1}). In image generation, some methods leverage pretrained priors to enhance the representation capacity of generative models, achieving promising generation performance \cite{li2023self, yu2024representation}. However, there are some limitations:

\begin{itemize}
	\item Requiring large-scale pretrained knowledge priors \cite{oquab2023dinov2}.
	\item Involving more costly two-stage training.
	
\end{itemize}

In this paper, we propose a Self-Conditional Representation Guidance (SCRG) that employs a Guidance Network (GN, $x_{\phi}$) to learn geometric features with reconstruction details from data distribution in an end-to-end manner. GN can provide adaptive perturbation and condition supervision signals for the Denoising Network (DN, $v_{\theta}$) to effectively learn geometric details,  while detached during inference.

Specifically, GN receives the multi-level perturbation features $F^{v_{\theta}}_{noise}$ with conditional guidance from DN, aligning real coordinates ($\bm{x_0}$) to reconstruct geometric details:

\vspace{-13pt}
\begin{equation}
	\begin{split}
		\label{f421}
		L(\phi) =||\bm{x_0} - \bm{x_\phi}(\bm{x_0},F^{v_{\theta}}_{noise})||^2.
	\end{split}
\end{equation}
\vspace{-13pt}

\begin{figure}[htp]
	\centering
	\includegraphics[width=0.48\textwidth]{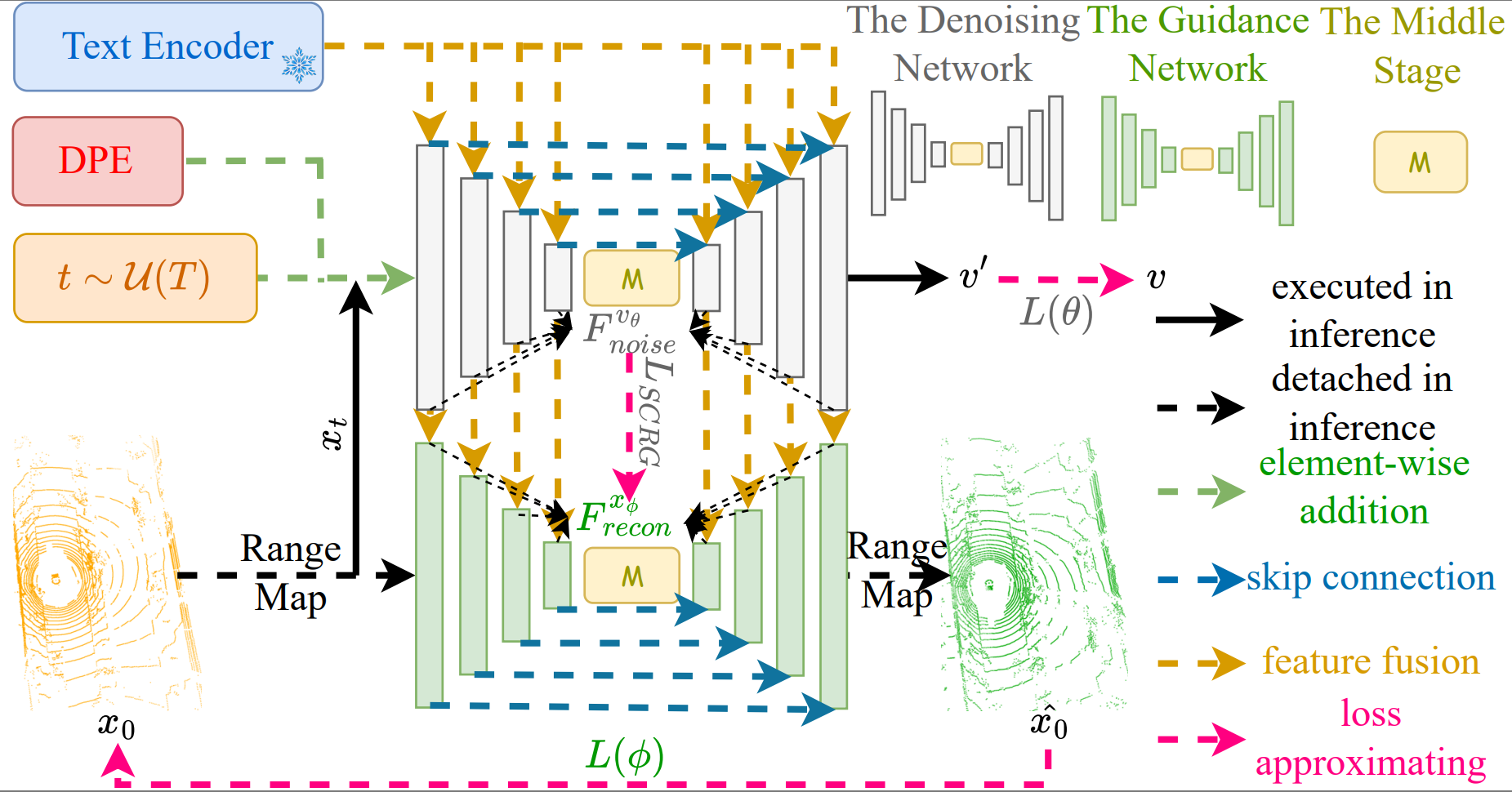}
	\vspace{-0.7cm}
	\caption{The overall framework of T2LDM. The Text Encoder (TE) encodes text prompts to generate semantically reliable features. Meanwhile, the Denoising Network (DN) models the denoising process under text guidance, DPE, and timestep. Furthermore, the Guidance Network (GN) introduces regularization with  reconstruction details for DN while detached during inference.}
	\label{fig3}
	
\end{figure}

Subsequently, to inject regularization for DN, $F^{v_{\theta}}_{noise}$ is aligned with the multi-scale reconstruction features $F^{x_{\phi}}_{recon}$ from GN (see Fig.~\ref{fig3} and Fig.~\ref{fig9}(bottom)):

\vspace{-12pt}
\begin{equation}
	\begin{split}
		\label{f422}
		L_{SCRG} =l_{recon}(F^{x_{\phi}}_{recon} - F^{v_{\theta}}_{noise}), 
	\end{split}
\end{equation}
\vspace{-12pt}

\noindent where $l_{recon}(\cdot)$ is a reconstruction loss (the cosine similarity in this paper, the additional ablation studies in \textcolor{blue}{SM}).

This simple and effective approach:
\begin{itemize}
	\item \textbf{With Lower Training Cost.} GN participates in gradient backpropagation only in the early stage and provides adaptive regularization to  DN in an end-to-end manner.
	\item \textbf{Without Inference Cost.} The detachable design of GN prevents cost and information leakage in inference.
	\item \textbf{With Faster Convergence.} The regularization from GN guides DN to learn high-frequency semantics for faster early-stage convergence (see Tab.~\ref{tab552} and Fig.~\ref{fig9}(bottom)).
	
\end{itemize}



\subsection{Directional Position Encoding}
\label{sec43}

RM represents the entire LiDAR scene by flattening the spherical projection \cite{milioto2019rangenet++}. However, window-based operations (e.g., convolution or local attention) perceive RM as a rectangular image rather than a circular image. This often leads to \textit{\textbf{directional confusion}}, making the model struggle to understand proper object orientations in the scene. The effect is most evident in distorted streets, as the starting angle is typically defined at the center of the street (see Fig.~\ref{fig2}).

In this paper, we design a Directional Position Encoding (DPE) for RM. By encoding the HFoV and VFoV angles, DPE can inject spherical geometric orientation priors, making the model perceive the true position of content in RM.

Specifically, given $\bm{x} \in \mathbb{R}^{b \times c \times h \times w}$ from $v_\theta$ or $x_{\phi}$, DPE first defines the angle of each pixel center in RM:

\vspace{-10pt}
\begin{equation}
	\begin{split}
		\label{f431}
		\bm{\theta} = 2\pi-(2\pi-0)*(w+0.5)/W, \quad\;\;\; \\
		\bm{\phi} = f_{up}-(f_{up}-f_{down})*(h+0.5)/H. 
	\end{split}
\end{equation}

Then, Fourier expansion and learnable gating are applied:


\vspace{-10pt}
\begin{equation}
	\begin{split}
		\label{f432}
		\mathrm{DPE}(\bm{\theta}, \bm{\phi}) = 
		Fourier^{K}(\bm{\theta}, \bm{\phi}), \\
		\bm{x'}=\bm{x}+\alpha*\mathrm{DPE}(\bm{\theta}, \bm{\phi}), \quad \;
	\end{split}
\end{equation}
where $\alpha$ means a learnable gating parameter. $K$ denotes the number of Fourier expansion terms, $Fourier^{K}(\bm{\theta}, \bm{\phi})=\bigoplus_{k=0}^{K-1}[\sin(2^{k}\bm{\theta}),\cos(2^{k}\bm{\theta}),\sin(2^{k}\bm{\phi}),\cos(2^{k}\bm{\phi})]$.

DPE encodes pixel angles via multi-level Fourier expansions to provide multi-scale directional priors in RM. Meanwhile, the learnable gating adaptively modulates the weights of multi-frequency features.

\subsection{Network Architecture}

In this section, we present T2LDM overall architecture, consisting of three key components in Fig.~\ref{fig3}: the Text Encoder (TE), the Denoising Network (DN), and the Guidance Network (GN) (parameters and optimizations in \textcolor{blue}{SM}).

\noindent\textbf{The Text Encoder.} TE encodes text prompts to provide semantic conditional features. The text prompt generalization largely depends on TE (see Sec.~\ref{sec33}). Therefore, we use the frozen CLIP to produce 768-dimensional text features with Text-to-Image semantic alignment \cite{ poole2022dreamfusion, lin2023magic3d, wu2024text2lidar}.

\noindent\textbf{The Denoising Network.} DN models the denoising, determining the generation results, following the U-Net architecture \cite{rombach2022high}. Each stage consists of Attention Block (AB) and Residual Block (RB) in the encoder and decoder. 

Specifically, AB receives text features $F^{CLIP}_{text}$ from TE as conditional guidance. This fuses the projected features $F^{v_{\theta}}_{noise} \in \mathbb{R}^{l \times C^{v_{\theta}}}$$\rightarrow$$(Q) \in \mathbb{R}^{l \times C}$ and $F^{CLIP}_{text} \in \mathbb{R}^{n \times 768}$$\rightarrow$$(K,V) \in \mathbb{R}^{n \times C}$ by a cross-attention block:

\vspace{-13pt}
\begin{equation}
	\begin{split}
		\label{f441}
		O = mlp(WV)+F^{v_{\theta}}_{noise}, \\
		F = ffn(O)+O, \quad\quad
	\end{split}
\end{equation}
\vspace{-5pt}

\noindent where $W \in \mathbb{R}^{l \times n} = softmax(\frac{QK^T}{\sqrt{C}})$ and $l=h \times w$.

\noindent \textit{\textbf{Replacing $F^{CLIP}_{text}$ with $F^{v_{\theta}}_{noise}$ means unconditional generation.}} The timestep $t$ and DPE are introduced into RB to identify denoising level and enhance scene fidelity.

\noindent\textbf{The Guidance Network.} GN produce the supervision signals with reconstruction details for DN (see Sec.~\ref{sec42}). This follows DN architecture with four stages in the encoder and decoder. To provide perturbation and conditional adaptation regularization, GN receives noise features from DN by Eq.~\ref{f441}. Meanwhile, we use only RM as input to ensure GN focuses on learning geometric features of data distribution.

\subsection{Training and Inference}

\noindent\textbf{Training.} As mentioned earlier (see Sec.~\ref{sec41} and Sec.~\ref{sec42}), T2LDM models the denoising process with SCRG. Therefore, the training objective is:

\vspace{-15pt}
\begin{equation}
	\begin{split}
		\label{f451}
		L_{total}=L(\theta)+L(\phi)+\lambda L_{SCRG},
	\end{split}
\end{equation}

\noindent where $\lambda$ means an epoch-wise weighting factor (details in \textcolor{blue}{SM}). Meanwhile, $x_{\phi}$ only participates in gradient backpropagation for the first 100K steps, then remains frozen.

\noindent\textbf{Inference.} T2LDM iteratively transforms $\bm{x_T}$ into $\bm{x'_0}$ by only $v_\theta$, due to the detachable design of $x_\phi$ (see Sec.~\ref{sec42}):

\vspace{-10pt}
\begin{equation}
	\begin{split}
		\bm{x_{t-1}} = \frac{1}{\sqrt{\alpha_t}} (\bm{x_t} - \frac{1-\alpha_t}{\sigma_t} [ \sigma_t \bm{x_t} + \sqrt{\bar{\alpha}_t} \, v_\theta] ) + \tilde{\sigma}_t \bm{\epsilon},
	\end{split}
\end{equation}
\vspace{-4pt}

\noindent where $\sigma_t=\sqrt{1-\bar{\alpha}_t}$, $\tilde{\sigma_t}=\sqrt{\frac{1 - \bar{\alpha}_{t-1}}{1 - \bar{\alpha}_t}(1-\alpha_t)}$. 

\begin{figure*}[htp]
	\centering
	\includegraphics[width=0.98\textwidth]{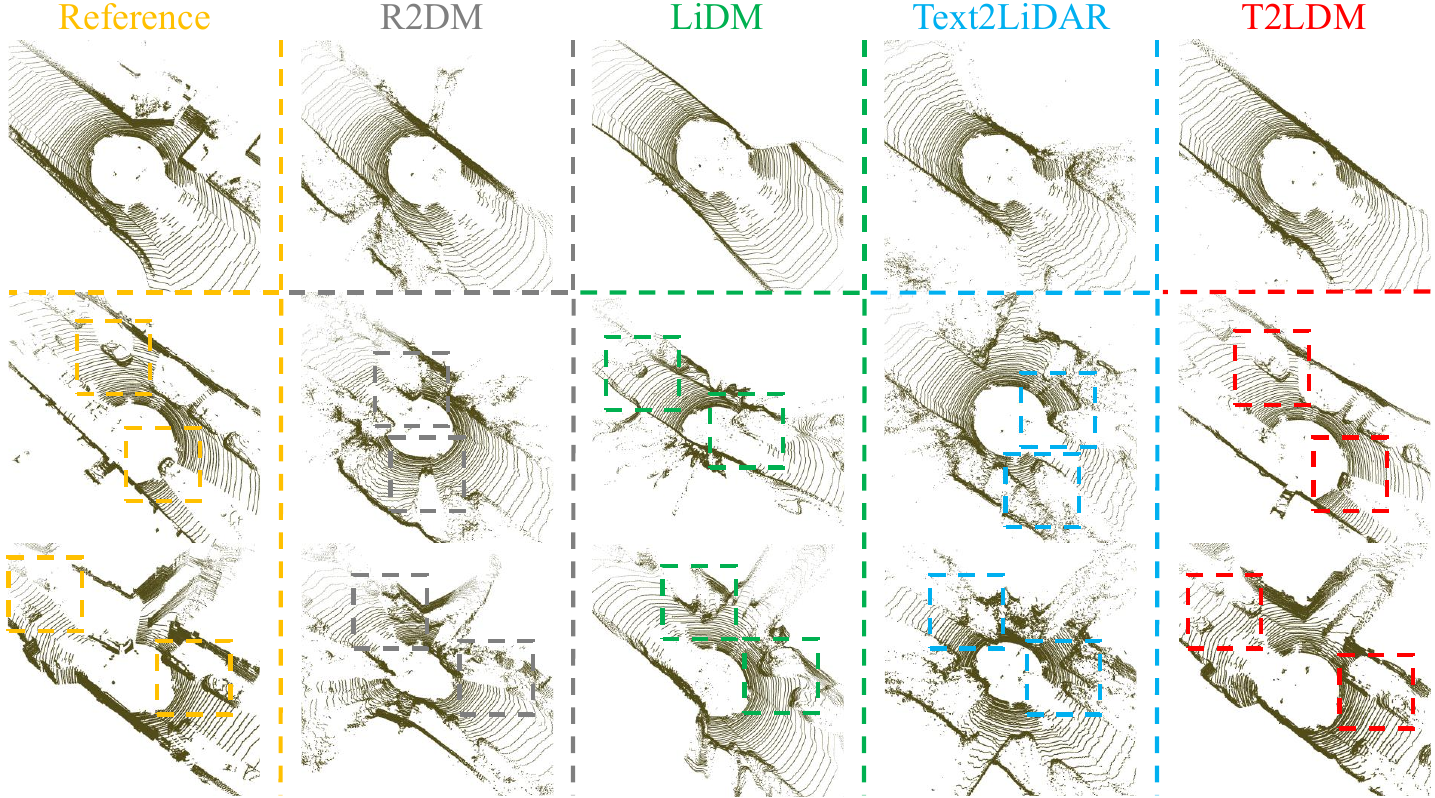}
	\vspace{-0.3cm}
	\caption{The generated visualization results on KITTI-360. Due to insufficient training priors, existing methods can only generate high-quality scenes with a few objects (top row). In contrast, T2LDM produces fine-grained geometric details even in complex multi-object scenes (bottom row). This is crucial for models to recognize 3D scenes in downstream tasks. For more visualizations, please refer to \textcolor{blue}{SM}.}
	\label{fig4}
	\vspace{-2mm}
\end{figure*}

\vspace{-3pt}
\section{Experiments}

\subsection{Experiment Setup}
\noindent\textbf{Dataset.} Two LiDAR benchmarks are used for training and evaluation: nuScenes \cite{caesar2020nuscenes} (32-beam, 34,149 samples) and KITTI-360 \cite{liao2022kitti} (64-beam, 76,165 samples). Meanwhile, the LiDAR data are projected into  $RM_{32beam}$ $\in \mathbb{R}^{32 \times 1024 \times 2}$ and $RM_{64beam}$ $\in \mathbb{R}^{64 \times 1024 \times 2}$ (see Sec.~\ref{sec41}).

\noindent\textbf{Metric.} FID (FSVD, FPVD), JSD, and MMD ($\times 10^{-4}$) are used for generation quality evaluation \cite{ran2024towards}. For fair comparison, we compute the true distribution over \textit{\textbf{all real samples}} instead of randomly selecting subsets for FID \cite{ran2024towards, wu2024text2lidar}. Meanwhile, for text-guided controllability, we propose TBR (see Sec.~\ref{sec31}), the matching rate between text semantics and 3D boxes obtained by a detector \cite{liu2025fshnet}. 

\vspace{-2pt}
\subsection{Unconditional Generation}
\vspace{-2pt}

\noindent\textbf{64-Beam LiDAR.} We first evaluate the generation quality on KITTI-360. Tab.~\ref{tab521} shows that T2LDM better matches the true distribution, as evidenced by the significantly lower FID. Benefiting from SCRG, T2LDM can more effectively learn realistic scene details from the data distribution. Therefore, compared with methods lacking sufficient prior training, T2LDM produces results with finer geometric details. Fig.~\ref{fig4} further presents the qualitative results.

\vspace{-5pt}
\begin{table}[h]
	\resizebox{0.475\textwidth}{!}{
			\begin{tabular}{p{3.0cm}|p{1.5cm}p{1.5cm}|p{1.1cm}p{1.1cm}p{1.2cm}p{1.1cm}}	
				\Xhline{1pt}
				
				{Methods}
				&\makecell[c]{Gen. Sam.}
				&\makecell[c]{Rea. Sam.}
				&\makecell[c]{FSVD$\downarrow$}
				&\makecell[c]{FPVD$\downarrow$}
				&\makecell[c]{JSD$\downarrow$}
				&\makecell[c]{MMD$\downarrow$}\\
				
				\Xhline{1pt}
				
				LiDARVAE \cite{caccia2019deep}
				&\makecell[c]{10000}
				&\makecell[c]{76165}      
				&\makecell[c]{281.14}
				&\makecell[c]{286.14}
				&\makecell[c]{0.35}
				&\makecell[c]{6.84}\\
				
				LiDARGAN \cite{caccia2019deep}
				&\makecell[c]{10000}
				&\makecell[c]{76165}       
				&\makecell[c]{346.23}
				&\makecell[c]{339.55}
				&\makecell[c]{0.38}
				&\makecell[c]{5.43}\\
				
				ProjectedGAN \cite{sauer2021projected}
				&\makecell[c]{10000}
				&\makecell[c]{76165}       
				&\makecell[c]{187.89}
				&\makecell[c]{201.62}
				&\makecell[c]{0.33}
				&\makecell[c]{3.45}\\
				
				LiDARGen \cite{zyrianov2022learning}
				&\makecell[c]{10000}
				&\makecell[c]{76165}       
				&\makecell[c]{238.72}
				&\makecell[c]{243.69}
				&\makecell[c]{0.32}
				&\makecell[c]{3.93}\\
				
				LiDM \cite{ran2024towards}
				&\makecell[c]{10000}
				&\makecell[c]{76165}       
				&\makecell[c]{211.68}
				&\makecell[c]{230.19}
				&\makecell[c]{0.35}
				&\makecell[c]{4.78}\\
				
				R2DM \cite{nakashima2024lidar}
				&\makecell[c]{10000}
				&\makecell[c]{76165}      
				&\makecell[c]{31.82}
				&\makecell[c]{35.94}
				&\makecell[c]{0.32}
				&\makecell[c]{4.05}\\
				
				Text2LiDAR \cite{wu2024text2lidar}
				&\makecell[c]{10000}
				&\makecell[c]{76165}      
				&\makecell[c]{51.55}
				&\makecell[c]{54.82}
				&\makecell[c]{0.33}
				&\makecell[c]{4.11}\\
				
				\cellcolor[rgb]{0.9725, 0.8078, 0.8} T2LDM 
				&\cellcolor[rgb]{0.9725, 0.8078, 0.8} \makecell[c]{10000}
				&\cellcolor[rgb]{0.9725, 0.8078, 0.8} \makecell[c]{76165}        
				&\cellcolor[rgb]{0.9725, 0.8078, 0.8} \makecell[c]{21.12}
				&\cellcolor[rgb]{0.9725, 0.8078, 0.8} \makecell[c]{25.39}
				&\cellcolor[rgb]{0.9725, 0.8078, 0.8} \makecell[c]{0.30}
				&\cellcolor[rgb]{0.9725, 0.8078, 0.8} \makecell[c]{3.35}\\
				
				\Xhline{1pt}
				
			\end{tabular}
		}
		\vspace{-5pt} 
		\caption{The results on KITTI-360. T2LDM significantly outperforms existing methods on all metrics.}
		\label{tab521}
		\vspace{-3mm}
	\end{table}
	
\begin{figure*}[htp]
	\centering
	\includegraphics[width=0.98\textwidth]{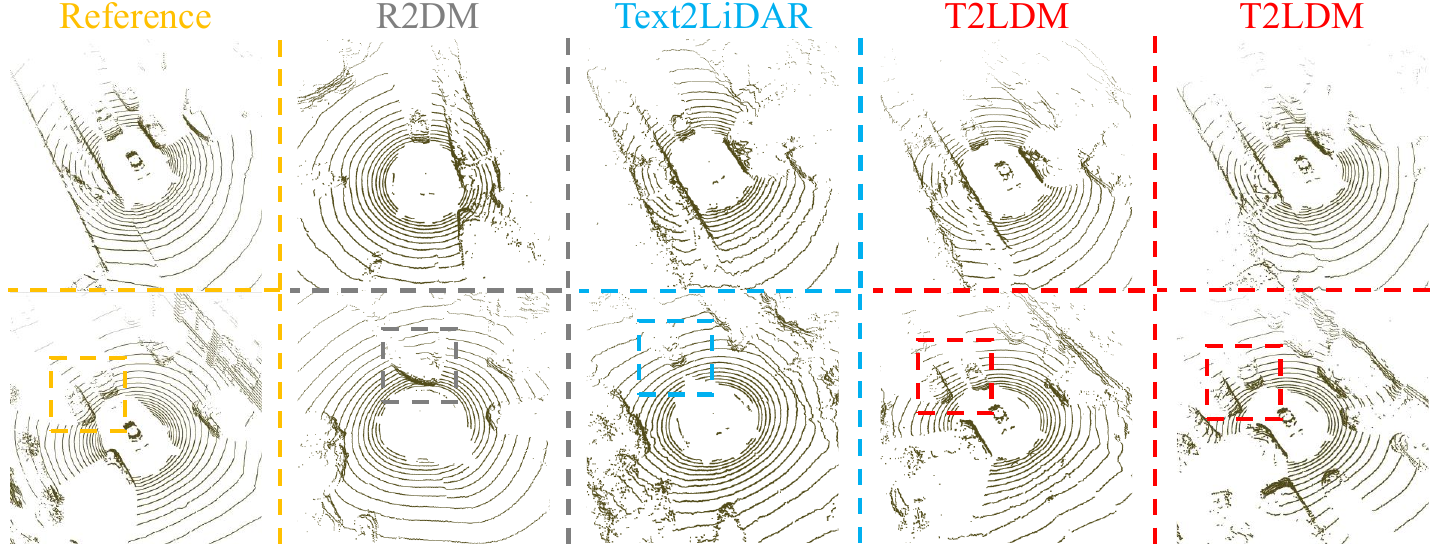}
	\vspace{-0.45cm}
	\caption{The generated visualization results on nuScenes. Similar to KITTI-360, existing methods can generate certain geometric details in scenes with few objects (top row) but struggle to handle complex multi-object scenes (bottom row), duo to the sufficient training data. This becomes more pronounced in the sparse scenes of nuScenes. In comparison, T2LDM can generate detailed objects even in multi-object scenes. Fig.~\ref{fig1} also shows that T2LDM can generate diverse structures for the same scene. More visualizations are provided in \textcolor{blue}{SM}.}
	\label{fig5}
	
\end{figure*}

\noindent\textbf{32-Beam LiDAR.} We also evaluate on the 32-beam benchmark. Compared with KITTI-360, nuScenes with fewer points and the larger spatial distance between points is more challenging due to the harder-to-capture geometric details. This also leads to poor performance for existing methods on nuScenes. However, T2LDM can achieve excellent generation results in Tab.~\ref{tab522}. As described in Sec.~\ref{sec42} and Sec.~\ref{sec43}, T2LDM captures effective geometric and directional priors during training through SCRG and DPE, enhancing the ability to perceive geometric details from the data distribution. Fig.~\ref{fig5} and Fig.~\ref{fig1} further demonstrates T2LDM can effectively generate rich and diverse details in sparse scenes. 
	
\vspace{-5pt}
\begin{table}[h]
	\resizebox{0.475\textwidth}{!}{
			\begin{tabular}{p{3.0cm}|p{1.5cm}p{1.5cm}|p{1.1cm}p{1.1cm}p{1.2cm}p{1.1cm}}	
				\Xhline{1pt}
				
				{Methods}
				&\makecell[c]{Gen. Sam.}
				&\makecell[c]{Rea. Sam.}
				&\makecell[c]{FSVD$\downarrow$}
				&\makecell[c]{FPVD$\downarrow$}
				&\makecell[c]{JSD$\downarrow$}
				&\makecell[c]{MMD$\downarrow$}\\
				
				\Xhline{1pt}
				
				R2DM \cite{nakashima2024lidar}
				&\makecell[c]{10000}
				&\makecell[c]{34149}      
				&\makecell[c]{86.54}
				&\makecell[c]{83.97}
				&\makecell[c]{0.42}
				&\makecell[c]{5.02}\\
				
				Text2LiDAR \cite{wu2024text2lidar}
				&\makecell[c]{10000}
				&\makecell[c]{34149}      
				&\makecell[c]{85.98}
				&\makecell[c]{80.94}
				&\makecell[c]{0.34}
				&\makecell[c]{3.45}\\
				
				\cellcolor[rgb]{1.0, 0.9490, 0.8} T2LDM 
				&\cellcolor[rgb]{1.0, 0.9490, 0.8} \makecell[c]{10000}
				&\cellcolor[rgb]{1.0, 0.9490, 0.8} \makecell[c]{34149}        
				&\cellcolor[rgb]{1.0, 0.9490, 0.8} \makecell[c]{64.21}
				&\cellcolor[rgb]{1.0, 0.9490, 0.8} \makecell[c]{62.85}
				&\cellcolor[rgb]{1.0, 0.9490, 0.8} \makecell[c]{0.26}
				&\cellcolor[rgb]{1.0, 0.9490, 0.8} \makecell[c]{3.01}\\
				
				\Xhline{1pt}
				
			\end{tabular}
		}
		\vspace{-5pt} 
		\caption{The results on nuScenes. T2LDM achieves superior generation results across all metrics in sparse scenes.}
		\label{tab522}
		\vspace{-1mm}
\end{table}
	
\vspace{-10pt}
\subsection{Text-Guided Generation}
Unlike other conditions with various constraints, text prompts are more accessible for human beings and can provide customized and diverse scene descriptions. 

We further validate the results on Text-to-LiDAR generation for T2LDM. Benefiting from detection priors, we can measure the generation controllability using the matching rate (TBK) between text prompts and 3D boxes obtained from a detector \cite{liu2025fshnet}. Meanwhile, as described in Sec.~\ref{sec34}, we perform evaluations on “weather, location”. Tab.~\ref{tab531} presents that T2LDM exhibits remarkable results in generation quality and controllability. With the perturbation and condition-guided adaptation regularization from GN, T2LDM can further perceive conditional features, improving the understanding of guided information. Furthermore, Fig.~\ref{fig6} shows the qualitative comparison.
	
\vspace{-7pt}
\begin{table}[h]
	\resizebox{0.475\textwidth}{!}{
			\begin{tabular}{p{3.0cm}|p{1.5cm}p{1.5cm}|p{1.1cm}p{1.1cm}p{1.2cm}p{1.1cm}p{1.5cm}}	
				\Xhline{1pt}
				
				{Methods}
				&\makecell[c]{Gen. Sam.}
				&\makecell[c]{Rea. Sam.}
				&\makecell[c]{FSVD$\downarrow$}
				&\makecell[c]{FPVD$\downarrow$}
				&\makecell[c]{JSD$\downarrow$}
				&\makecell[c]{MMD$\downarrow$}
				&\makecell[c]{TBK$(\%)$$\uparrow$}\\
				
				\Xhline{1pt}
				
				R2DM \cite{nakashima2024lidar}
				&\makecell[c]{10000}
				&\makecell[c]{34149}      
				&\makecell[c]{91.15}
				&\makecell[c]{88.55}
				&\makecell[c]{0.45}
				&\makecell[c]{5.11}
				&\makecell[c]{15.45}\\
				
				Text2LiDAR \cite{wu2024text2lidar}
				&\makecell[c]{10000}
				&\makecell[c]{34149}      
				&\makecell[c]{90.13}
				&\makecell[c]{87.62}
				&\makecell[c]{0.38}
				&\makecell[c]{4.01}
				&\makecell[c]{17.15}\\
				
				\cellcolor[rgb]{0.8352, 0.9098, 0.8314} T2LDM 
				&\cellcolor[rgb]{0.8352, 0.9098, 0.8314} \makecell[c]{10000}
				&\cellcolor[rgb]{0.8352, 0.9098, 0.8314} \makecell[c]{34149}        
				&\cellcolor[rgb]{0.8352, 0.9098, 0.8314} \makecell[c]{66.93}
				&\cellcolor[rgb]{0.8352, 0.9098, 0.8314} \makecell[c]{65.84}
				&\cellcolor[rgb]{0.8352, 0.9098, 0.8314} \makecell[c]{0.28}
				&\cellcolor[rgb]{0.8352, 0.9098, 0.8314} \makecell[c]{3.05}
				&\cellcolor[rgb]{0.8352, 0.9098, 0.8314} \makecell[c]{23.44}
				\\
				
				\Xhline{1pt}
				
			\end{tabular}
		}
		\vspace{-5pt} 
		\caption{The text-guided results on nuScenes. T2LDM exhibits outstanding performance in generation quality and controllability.}
		\label{tab531}
		\vspace{-1mm}
\end{table}
	
\vspace{-14pt}
\subsection{Other Conditional Generation}
\vspace{-4pt}
	
By freezing unconditional DN, T2LDM can achieve various conditional tasks. \textit{\textbf{This also marks the first exploration of ControlNet \cite{zhang2023adding} into 3D generation in non-latent DDPMs}} (please refer to the implementation details in \textcolor{blue}{SM}). 
	
\vspace{-8pt}
\begin{table}[h]
	\scriptsize
	\resizebox{0.48\textwidth}{!}{
		\begin{tabular}{p{1.5cm}p{0.6cm}p{0.8cm}p{0.8cm}p{0.005cm}p{0.6cm}p{0.8cm}p{0.8cm}}	
			\Xhline{1pt}
			
			\makecell[l]{\multirow{2}{*}{Methods}}
			&\multicolumn{3}{c}{$4\times$} 
			&\quad
			&\multicolumn{3}{c}{$8\times$} \\
			\cline{2-4} \cline{6-8}
			
			&\makecell[c]{CD$\downarrow$}
			&\makecell[c]{MSE$\downarrow$}
			&\makecell[c]{EMD$\downarrow$}
			&\quad
			&\makecell[c]{CD$\downarrow$}
			&\makecell[c]{MSE$\downarrow$}
			&\makecell[c]{EMD$\downarrow$}\\
			\hline
			
			\makecell[l]{Grad-PU \cite{he2023grad}}
			&\makecell[c]{0.400}
			&\makecell[c]{4.169}
			&\makecell[c]{2.324}
			&\quad
			&\makecell[c]{0.364}
			&\makecell[c]{4.031}
			&\makecell[c]{2.142}\\
			
			\makecell[l]{PUDM \cite{qu2024conditional}}
			&\makecell[c]{0.198}
			&\makecell[c]{4.275}
			&\makecell[c]{2.124}
			&\quad
			&\makecell[c]{0.103}
			&\makecell[c]{4.102}
			&\makecell[c]{1.914}\\
			
			\cellcolor[rgb]{1.0, 0.9020, 0.8} \makecell[l]{T2LDM}
			&\cellcolor[rgb]{1.0, 0.9020, 0.8} \makecell[c]{0.104}
			&\cellcolor[rgb]{1.0, 0.9020, 0.8} \makecell[c]{3.610}
			&\cellcolor[rgb]{1.0, 0.9020, 0.8} \makecell[c]{1.987}
			&\cellcolor[rgb]{1.0, 0.9020, 0.8} \quad
			&\cellcolor[rgb]{1.0, 0.9020, 0.8} \makecell[c]{0.074}
			&\cellcolor[rgb]{1.0, 0.9020, 0.8} \makecell[c]{3.574}
			&\cellcolor[rgb]{1.0, 0.9020, 0.8} \makecell[c]{1.910}\\
			\Xhline{1pt}
			
		\end{tabular}
	}
	\vspace{-5pt} 
	\caption{The results of the $4 \times$ rate and the $8 \times$ rate on nuScenes. T2LDM exhibits significantly upsampling results.}
	\label{tab541}
\end{table}
		
\begin{figure}[htp]
	\centering
	\includegraphics[width=0.48\textwidth]{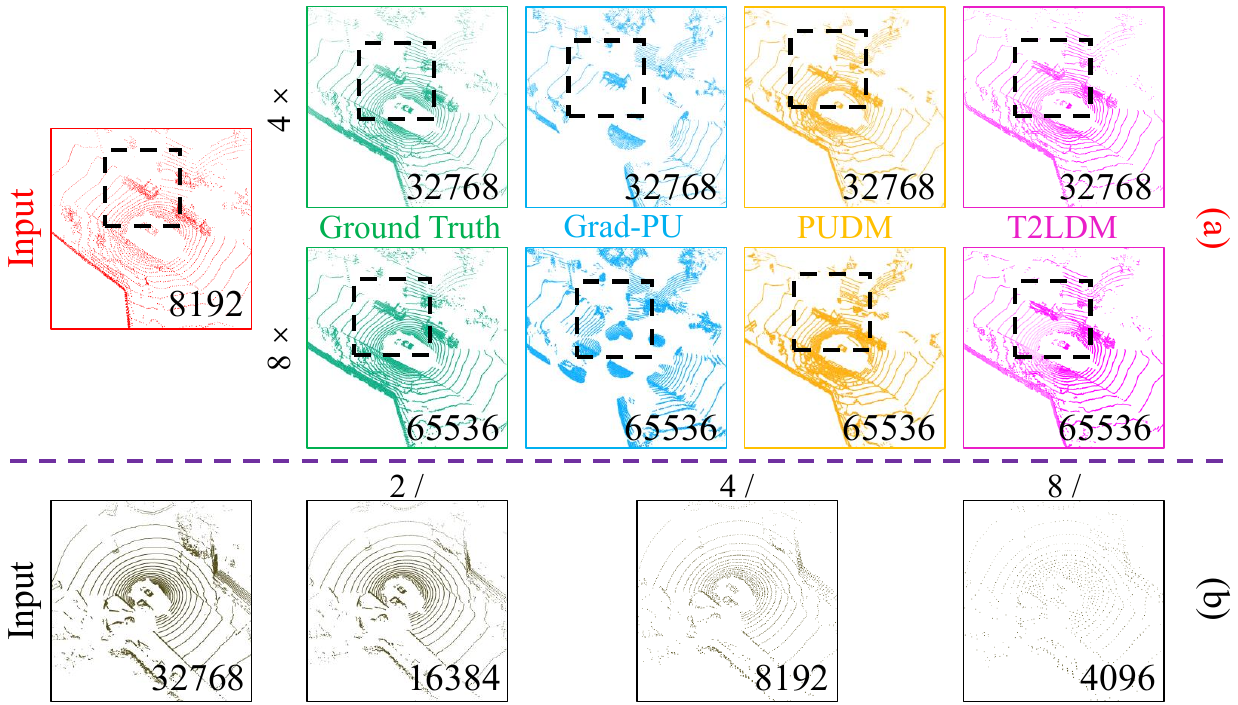}
	\vspace{-0.7cm}
	\caption{(a) Qualitative results of upsampling. (b) Without retraining, T2LDM can perform downsampling at arbitrary rates.}
	\label{fig7}
	\vspace{-0.6cm}
\end{figure}

\vspace{-13pt} 
\noindent\textbf{Sparse-to-Dense Generation.} We downsample the training set (28,140 samples) by $4 \times$ using FPS as sparse LiDAR for training, while the original and $2 \times$ upsampled validation sets (6019 samples) are used as $4 \times$ and $8 \times$ Ground Truth for evaluation. We follow existing methods \cite{he2023grad, qu2024conditional} by directly validating the PU-GAN \cite{li2019pu} pretrained model on nuScenes. Meanwhile, we find that PUDM shows better qualitative but inconsistent quantitative results than Grad-PU. For fair comparison, we normalize point coordinates to [0,1] (CD$\times$$10^{-5}$, MSE$\times$$10^{-5}$, EMD$\times$$10^{-3}$). Tab.~\ref{tab541} presents the remarkable upsampling results for T2LDM. Fig.~\ref{fig7}(a) further illustrates the superior qualitative results.

\vspace{-8pt}

\noindent\textbf{Dense-to-Sparse Generation.} Meanwhile, since the output LiDAR data shape is determined by the input noise size, we can directly achieve downsampling by upsampling-enabled T2LDM. Fig.~\ref{fig7}(b) shows the Dense-to-Sparse results.

\begin{figure}[htp]
	\centering
	\includegraphics[width=0.48\textwidth]{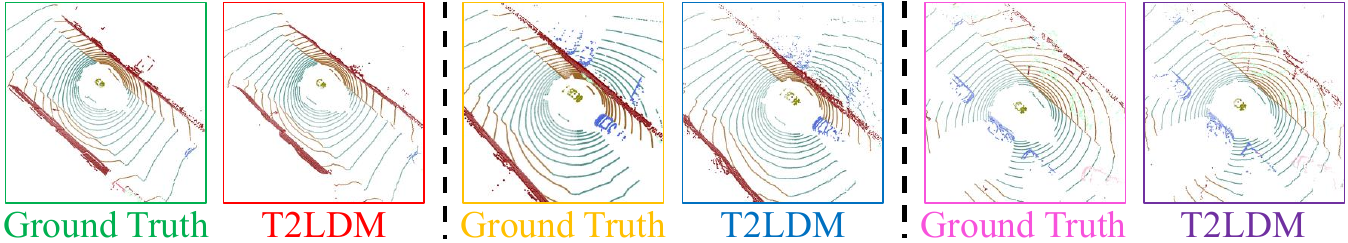}
	\vspace{-0.7cm}
	\caption{Semantic-to-Lidar results on nuScenes. With  the DN frozen, T2LDM shows remarkable semantic-guided generation.}
	\label{fig8}
	
\end{figure}

\begin{figure*}[htp]
	\centering
	\includegraphics[width=0.98\textwidth]{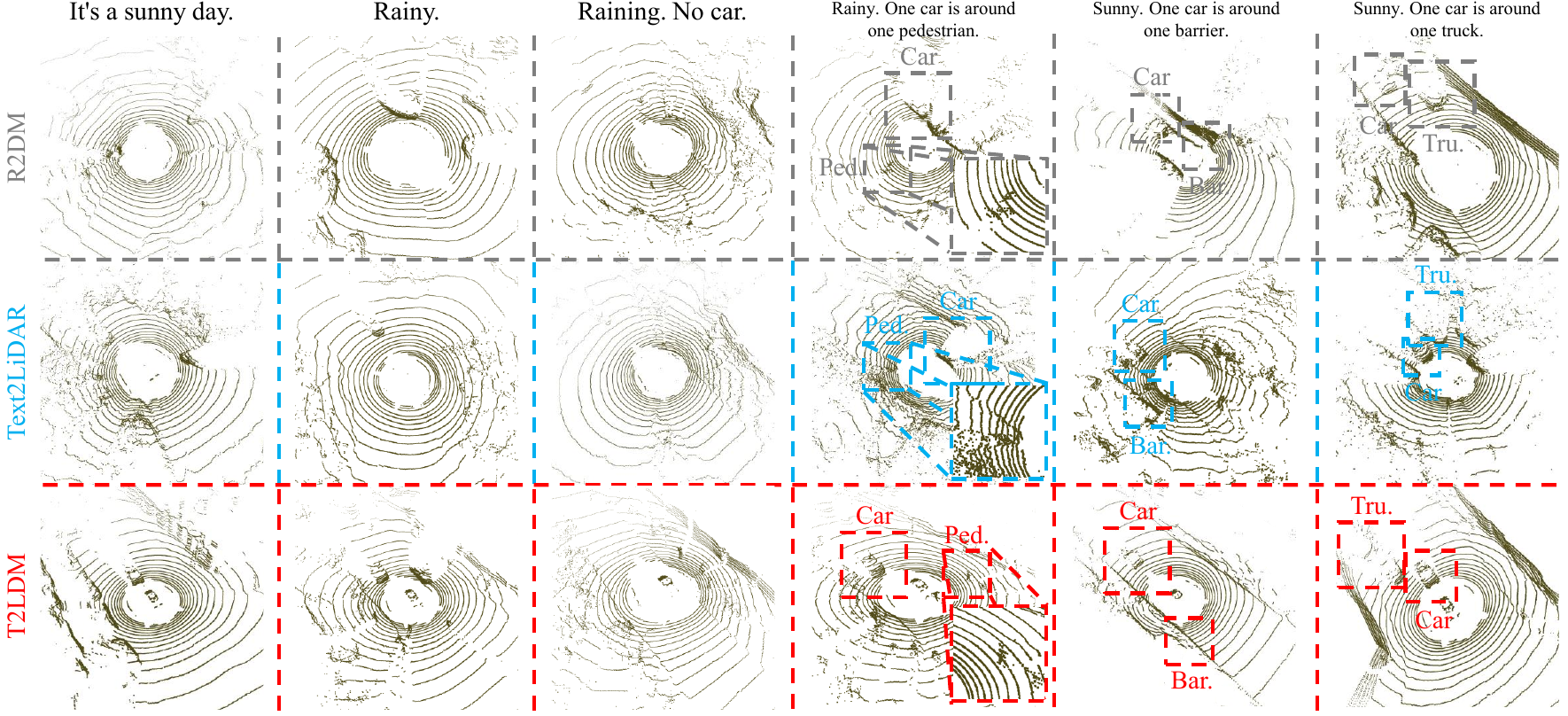}
	\vspace{-0.3cm}
	\caption{Text-guided generation results on nuScenes. Existing methods produce overly smooth results with insufficient object details, making the difficult to satisfy the text semantics. However, T2LDM shows superior detail generation that aligns well with the text prompts.}
	\label{fig6}
	
\end{figure*}

\noindent\textbf{Semantic-to-LiDAR Generation.} Furthermore, we also implement Semantic-to-LiDAR generation on nuScenes. Tab.~\ref{tab542} and Fig.~\ref{fig8} present the quantitative and qualitative results, respectively (SemanticKITTI \cite{behley2019semantickitti} results in \textcolor{blue}{SM}).

\vspace{-7pt}
\begin{table}[h]
	\resizebox{0.475\textwidth}{!}{
			\begin{tabular}{p{3.0cm}|p{1.5cm}p{1.5cm}|p{1.1cm}p{1.1cm}p{1.2cm}p{1.1cm}}	
				\Xhline{1pt}
				
				{Methods}
				&\makecell[c]{Gen. Sam.}
				&\makecell[c]{Rea. Sam.}
				&\makecell[c]{FSVD$\downarrow$}
				&\makecell[c]{FPVD$\downarrow$}
				&\makecell[c]{JSD$\downarrow$}
				&\makecell[c]{MMD$\downarrow$}\\
				
				\Xhline{1pt}
				
				T2LDM+Uncon. 
				&\makecell[c]{10000}
				&\makecell[c]{34149}        
				&\makecell[c]{64.21}
				&\makecell[c]{62.85}
				&\makecell[c]{0.26}
				&\makecell[c]{3.01}
				\\
				
				\cellcolor[rgb]{0.8549, 0.9098, 0.9882} T2LDM+Seman. 
				&\cellcolor[rgb]{0.8549, 0.9098, 0.9882} \makecell[c]{10000}
				&\cellcolor[rgb]{0.8549, 0.9098, 0.9882} \makecell[c]{34149}        
				&\cellcolor[rgb]{0.8549, 0.9098, 0.9882} \makecell[c]{62.91}
				&\cellcolor[rgb]{0.8549, 0.9098, 0.9882} \makecell[c]{60.54}
				&\cellcolor[rgb]{0.8549, 0.9098, 0.9882} \makecell[c]{0.23}
				&\cellcolor[rgb]{0.8549, 0.9098, 0.9882} \makecell[c]{2.94}
				\\
				
				\Xhline{1pt}
				
			\end{tabular}
		}
		\vspace{-5pt} 
		\caption{The Semantic-to-LiDAR results on nuScenes. T2LDM achieves excellent results for semantic map guidance generation.}
		\label{tab542}
		\vspace{-1mm}
\end{table}
	
\vspace{-15pt}
	
\subsection{Ablation Study} 
\noindent\textbf{Component Effectiveness.} We first perform ablations for component effectiveness. Tab.~\ref{tab551} shows that removing SCRG and DPE leads to a significant drop for the generation quality and controllability of T2LDM. As discussed in Sec.~\ref{sec42} and Sec.~\ref{sec43}, insufficient training priors make learning real details from the data distribution difficult for the model, leading to blurry objects in the generated scenes.
	
\vspace{-7pt}
\begin{table}[h]
	\resizebox{0.475\textwidth}{!}{
			\begin{tabular}{p{3.0cm}|p{1.5cm}p{1.5cm}|p{1.1cm}p{1.1cm}p{1.2cm}p{1.1cm}p{1.5cm}}	
				\Xhline{1pt}
				
				{Methods}
				&\makecell[c]{Gen. Sam.}
				&\makecell[c]{Rea. Sam.}
				&\makecell[c]{FSVD$\downarrow$}
				&\makecell[c]{FPVD$\downarrow$}
				&\makecell[c]{JSD$\downarrow$}
				&\makecell[c]{MMD$\downarrow$}
				&\makecell[c]{TBK$(\%)$$\uparrow$}\\
				
				\Xhline{1pt}
				
				T2LDM$^{\emptyset}$
				&\makecell[c]{10000}
				&\makecell[c]{34149}      
				&\makecell[c]{73.64}
				&\makecell[c]{71.91}
				&\makecell[c]{0.34}
				&\makecell[c]{3.21}
				&\makecell[c]{19.32}\\
				
				T2LDM$^{D}$ 
				&\makecell[c]{10000}
				&\makecell[c]{34149}      
				&\makecell[c]{71.32}
				&\makecell[c]{70.44}
				&\makecell[c]{0.32}
				&\makecell[c]{3.15}
				&\makecell[c]{20.95}\\
				
				T2LDM$^{S}$ 
				&\makecell[c]{10000}
				&\makecell[c]{34149}      
				&\makecell[c]{68.45}
				&\makecell[c]{67.77}
				&\makecell[c]{0.30}
				&\makecell[c]{3.07}
				&\makecell[c]{22.15}\\
				
				\cellcolor[rgb]{0.9922, 0.8275, 0.9647} T2LDM 
				&\cellcolor[rgb]{0.9922, 0.8275, 0.9647} \makecell[c]{10000}
				&\cellcolor[rgb]{0.9922, 0.8275, 0.9647} \makecell[c]{34149}        
				&\cellcolor[rgb]{0.9922, 0.8275, 0.9647} \makecell[c]{66.93}
				&\cellcolor[rgb]{0.9922, 0.8275, 0.9647} \makecell[c]{65.84}
				&\cellcolor[rgb]{0.9922, 0.8275, 0.9647} \makecell[c]{0.28}
				&\cellcolor[rgb]{0.9922, 0.8275, 0.9647} \makecell[c]{3.05}
				&\cellcolor[rgb]{0.9922, 0.8275, 0.9647} \makecell[c]{23.44}
				\\
				
				\Xhline{1pt}
				
			\end{tabular}
		}
		\vspace{-5pt} 
		\caption{Ablation study of component effectiveness for text-guided generation on nuScenes. T2LDM$^{\emptyset}$, T2LDM$^{D}$, and T2LDM$^{S}$ denote removing DPE and SCRG, keeping only DPE, and keeping only SCRG, respectively. DPE and SCRG can provide effective priors and regularization, enhancing scene fidelity.}
		\label{tab551}
		\vspace{-3mm}
\end{table}
		
\begin{figure}[htp]
	\centering
	\includegraphics[width=0.48\textwidth]{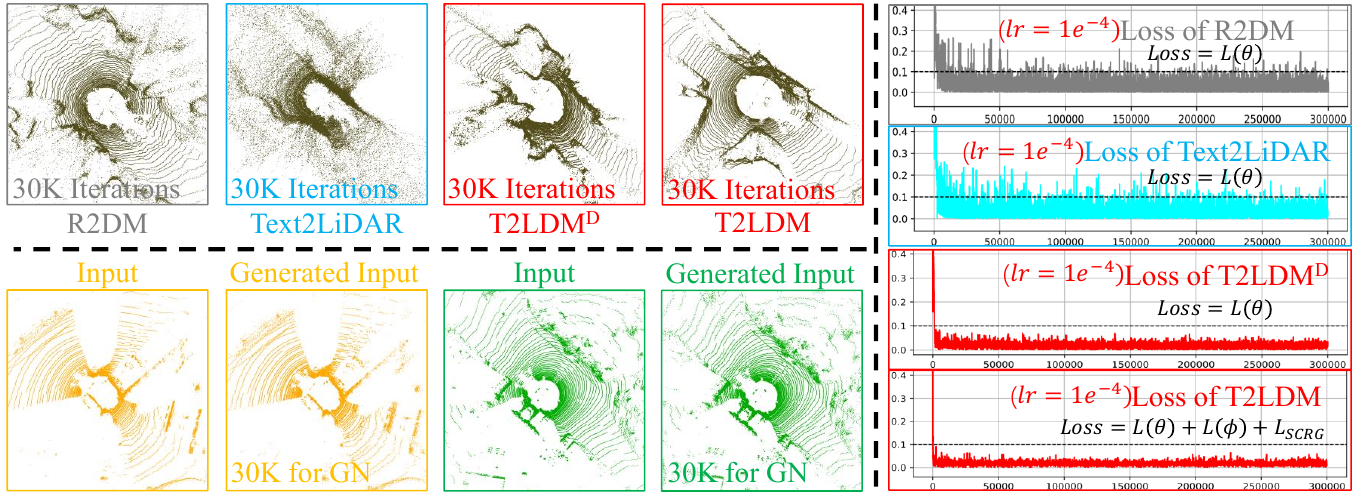}
	\vspace{-0.7cm}
	\caption{Ablation study of convergence speed on KITTI360. The loss combination of T2LDM shows more stable and superior.}
	\label{fig9}
	\vspace{-6mm}
\end{figure}

\noindent\textbf{Convergence Speed.} We further evaluate the effect of SCRG on the convergence speed for T2LDM. Fig.~\ref{fig9}(top) shows that SCRG allows DN to capture high frequency details early in training, generating detailed scene structures. Fig.~\ref{fig9}(bottom) presents that GN can learn rich geometric detail features to offer effective regularization. Fig.~\ref{fig9}(right) shows that SCRG enables faster and more stable convergence of T2LDM. Tab.~\ref{tab552} presents results at 30k iterations.

				
\vspace{-7pt}
\begin{table}[h]
	\resizebox{0.475\textwidth}{!}{
			\begin{tabular}{p{3.0cm}|p{1.8cm}p{1.5cm}p{1.5cm}|p{1.1cm}p{1.1cm}p{1.2cm}p{1.1cm}}	
				\Xhline{1pt}
				
				{Methods (30k Itera.)}
				&\makecell[c]{Inf. Param.}
				&\makecell[c]{Inf. Steps}
				&\makecell[c]{Gen. Sam.}
				&\makecell[c]{FSVD$\downarrow$}
				&\makecell[c]{FPVD$\downarrow$}
				&\makecell[c]{JSD$\downarrow$}
				&\makecell[c]{MMD$\downarrow$}\\
				
				\Xhline{1pt}
				
				R2DM \cite{nakashima2024lidar}
				&\makecell[c]{31.1M}
				&\makecell[c]{1024}
				&\makecell[c]{10000}      
				&\makecell[c]{175.82}
				&\makecell[c]{152.57}
				&\makecell[c]{0.55}
				&\makecell[c]{10.08}\\
				
				Text2LiDAR \cite{wu2024text2lidar}
				&\makecell[c]{45.8M}  
				&\makecell[c]{1024}
				&\makecell[c]{10000}      
				&\makecell[c]{340.95}
				&\makecell[c]{320.43}
				&\makecell[c]{0.84}
				&\makecell[c]{16.61}\\
				
				T2LDM$^{D}$
				&\makecell[c]{30.4M}  
				&\makecell[c]{1024}
				&\makecell[c]{10000}      
				&\makecell[c]{91.32}
				&\makecell[c]{88.44}
				&\makecell[c]{0.45}
				&\makecell[c]{1.45}\\
				
				\cellcolor[rgb]{0.8824, 0.8353, 0.9059} T2LDM 
				&\cellcolor[rgb]{0.8824, 0.8353, 0.9059} \makecell[c]{30.4M}
				&\cellcolor[rgb]{0.8824, 0.8353, 0.9059} \makecell[c]{1024}
				&\cellcolor[rgb]{0.8824, 0.8353, 0.9059} \makecell[c]{10000}        
				&\cellcolor[rgb]{0.8824, 0.8353, 0.9059} \makecell[c]{47.29}
				&\cellcolor[rgb]{0.8824, 0.8353, 0.9059} \makecell[c]{55.57}
				&\cellcolor[rgb]{0.8824, 0.8353, 0.9059} \makecell[c]{0.35}
				&\cellcolor[rgb]{0.8824, 0.8353, 0.9059} \makecell[c]{0.55}\\
				
				\Xhline{1pt}
				
			\end{tabular}
		}
		\vspace{-5pt} 
		\caption{The results on KITTI-360 at 30k iterations. SCRG enables T2LDM to learn high-frequency semantics early.}
		\label{tab552}
		\vspace{-3mm}
\end{table}

\noindent\textbf{End-to-End vs. Pretrained Mode.} We also conduct the ablation study on the end-to-end and pretrained paradigms of SCRG. As shown in Tab.~\ref{tab553}, pretrained training even leads to performance degradation, since GN cannot perceive DN features to provide adaptive regularization \cite{leng2025repae}. In contrast, end-to-end paradigm enables joint training of GN and DN, allowing GN to produce feature-aware supervision signals.

\vspace{-7pt}
\begin{table}[h]
	\resizebox{0.475\textwidth}{!}{
			\begin{tabular}{p{3.0cm}|p{1.5cm}p{1.5cm}|p{1.1cm}p{1.1cm}p{1.2cm}p{1.1cm}}	
				\Xhline{1pt}
				
				{Methods}
				&\makecell[c]{Gen. Sam.}
				&\makecell[c]{Rea. Sam.}
				&\makecell[c]{FSVD$\downarrow$}
				&\makecell[c]{FPVD$\downarrow$}
				&\makecell[c]{JSD$\downarrow$}
				&\makecell[c]{MMD$\downarrow$}\\
				
				\Xhline{1pt}
				
				T2LDM$^{D}$ 
				&\makecell[c]{10000}
				&\makecell[c]{34149}      
				&\makecell[c]{68.11}
				&\makecell[c]{67.32}
				&\makecell[c]{0.31}
				&\makecell[c]{3.11}\\
				
				Pretrained Mode
				&\makecell[c]{10000}
				&\makecell[c]{34149}      
				&\makecell[c]{68.77}
				&\makecell[c]{67.94}
				&\makecell[c]{0.33}
				&\makecell[c]{3.12}\\
				
				\cellcolor[rgb]{0.7800, 0.7800, 0.7800} End-to-End Mode 
				&\cellcolor[rgb]{0.7800, 0.7800, 0.7800} \makecell[c]{10000}
				&\cellcolor[rgb]{0.7800, 0.7800, 0.7800} \makecell[c]{34149}        
				&\cellcolor[rgb]{0.7800, 0.7800, 0.7800} \makecell[c]{64.21}
				&\cellcolor[rgb]{0.7800, 0.7800, 0.7800} \makecell[c]{62.85}
				&\cellcolor[rgb]{0.7800, 0.7800, 0.7800} \makecell[c]{0.26}
				&\cellcolor[rgb]{0.7800, 0.7800, 0.7800} \makecell[c]{3.01}\\
				
				\Xhline{1pt}
				
			\end{tabular}
		}
		\vspace{-5pt} 
		\caption{Ablation study of end-to-end vs. pretrained training for SCRG on nuScenes. The end-to-end mode yields better results.}
		\label{tab553}
		\vspace{-7mm}
\end{table}

\section{Conclusion}
\vspace{-5pt}

In this paper, we proposed a Text-to-LiDAR diffusion model that leverages a self-conditioned representation guidance to enhance details in generated LiDAR scenes. Meanwhile, a directional position prior is used to resolve directional confusion, correcting road distortion. Moreover, we design a 3D box-based annotation scheme to construct a content-composable Text-LiDAR benchmark, offering a controllability generation metric and insights to encourage researchers to focus on Text-to-LiDAR generation.

\newpage


\section*{Supplementary Material}

\setcounter{equation}{0}
\setcounter{section}{0}
\setcounter{figure}{0}
\setcounter{table}{0}

\vspace{4mm}

Due to space limitations in the Main Text (\textcolor{orange}{MT}), we include additional ablation and comparison experiments, derivations, benchmark details, implementations, and discussions in the supplementary material. We first present additional ablation studies (Sec.~\ref{supp_sec1}) and comparison experiments (Sec.~\ref{supp_sec2}). Then, we logically derive the conversion relationships among $\bm{v}$, $\bm{x_0}$ and $\bm{\epsilon}$  (Sec.~\ref{supp_sec3}) and provide the Text-LiDAR benchmark processing (Sec.~\ref{supp_sec4}). Next, the implementation and optimization details (Sec.~\ref{supp_sec5}) of T2LDM are presented. Finally, we discuss the limitations/future works (Sec.~\ref{supp_sec6}) and visualize additional results (Sec.~\ref{supp_sec7}).

\section{Additional Ablation Study}
\label{supp_sec1}

\subsection{Reconstruction Loss of SCRG}

As discussed in Sec.~4.2 of \textcolor{orange}{MT}, SCRG can employ arbitrary reconstruction losses to align the noise feature $F^{v_{\theta}}_{noise}$ from \textcolor{gray}{DN} with the reconstructed feature $F^{x_{\phi}}_{recon}$ from \textcolor[rgb]{0,0.6902,0.3137}{GN}, enhancing the geometric detail perception of $F^{v_{\theta}}_{noise}$. In Tab.~\ref{supp_tab111}, cosine similarity demonstrates better generation performance. This is because unlike MSE ($\mathcal{L}_1$ or $\mathcal{L}_2$), cosine similarity focuses on \textit{\textbf{directional consistency}} rather than numerical scale consistency. In fact, \textit{\textbf{semantic information is determined by the feature direction rather than the feature magnitude}} (semantic similarity emerges in the angular geometry of the embedding space \cite{radford2021learning}). For example, enlarging or shrinking a feature vector does not affect the semantic meaning. This also aligns with findings from prior representation learning studies \cite{radford2021learning, yu2024representation}.

\vspace{-5pt}
\begin{table}[h]
	\resizebox{0.475\textwidth}{!}{
			\begin{tabular}{p{3.0cm}|p{1.5cm}p{1.5cm}|p{1.1cm}p{1.1cm}p{1.2cm}p{1.1cm}}	
				\Xhline{1pt}
				
				{Loss}
				&\makecell[c]{Gen. Sam.}
				&\makecell[c]{Rea. Sam.}
				&\makecell[c]{FSVD$\downarrow$}
				&\makecell[c]{FPVD$\downarrow$}
				&\makecell[c]{JSD$\downarrow$}
				&\makecell[c]{MMD$\downarrow$}\\
				
				\Xhline{1pt}
				
				$\mathcal{L}_1$ Loss
				&\makecell[c]{10000}
				&\makecell[c]{34149}      
				&\makecell[c]{66.12}
				&\makecell[c]{64.25}
				&\makecell[c]{0.29}
				&\makecell[c]{3.05}\\
				
				MSE Loss
				&\makecell[c]{10000}
				&\makecell[c]{34149}      
				&\makecell[c]{65.45}
				&\makecell[c]{63.52}
				&\makecell[c]{0.28}
				&\makecell[c]{3.03}\\
				
				\cellcolor[rgb]{0.9725, 0.8078, 0.8} Cos. Sim. 
				&\cellcolor[rgb]{0.9725, 0.8078, 0.8} \makecell[c]{10000}
				&\cellcolor[rgb]{0.9725, 0.8078, 0.8} \makecell[c]{34149}        
				&\cellcolor[rgb]{0.9725, 0.8078, 0.8} \makecell[c]{64.21}
				&\cellcolor[rgb]{0.9725, 0.8078, 0.8} \makecell[c]{62.85}
				&\cellcolor[rgb]{0.9725, 0.8078, 0.8} \makecell[c]{0.26}
				&\cellcolor[rgb]{0.9725, 0.8078, 0.8} \makecell[c]{3.01}\\
				
				\Xhline{1pt}
				
			\end{tabular}
		}
		\vspace{-5pt} 
		\caption{The results on nuScenes. Cosine similarity achieves superior generation results than other reconstruction losses.}
		\label{supp_tab111}
		\vspace{-4mm}
\end{table}
	
\subsection{Reconstruction Loss of \textcolor{gray}{DN}}

We also investigate the choice of reconstruction loss for \textcolor{gray}{DN}. As presented in Tab.~\ref{supp_tab121}, Huber loss demonstrates superior performance. Since the target $\bm{v}$ is formed by combining $\bm{x_0}$ and $\bm{\epsilon}$, which exhibits highly the distribution difference. The distribution is inherently sharper and more susceptible to large deviations than that of $\bm{\epsilon}$. The quadratic penalty of $\mathcal{L}_2$ magnifies outliers in high-noise stages, which ultimately results in blurrier generation. Meanwhile, although the linear penalty renders $\mathcal{L}_1$ robust against outliers, the gradient discontinuity often results in unstable training behavior. In contrast, The Huber loss transitions to $\mathcal{L}_2$ loss in regions of low error and to $\mathcal{L}_1$ loss in regions of high error, making the more robust to outliers while maintaining continuous gradients. Therefore, \textit{\textbf{Huber loss is often more suitable for diffusion models targeting $\bm{v}$ in LiDAR generation.}}
	
\begin{table}[h]
	\vspace{1.2cm}
	\resizebox{0.475\textwidth}{!}{
		\begin{tabular}{p{3.0cm}|p{1.5cm}p{1.5cm}|p{1.1cm}p{1.1cm}p{1.2cm}p{1.1cm}}	
			\Xhline{1pt}
			
			{Methods}
			&\makecell[c]{Gen. Sam.}
			&\makecell[c]{Rea. Sam.}
			&\makecell[c]{FSVD$\downarrow$}
			&\makecell[c]{FPVD$\downarrow$}
			&\makecell[c]{JSD$\downarrow$}
			&\makecell[c]{MMD$\downarrow$}\\
			
			\Xhline{1pt}
			
			$\mathcal{L}_1$ Loss
			&\makecell[c]{10000}
			&\makecell[c]{76165}      
			&\makecell[c]{23.14}
			&\makecell[c]{27.55}
			&\makecell[c]{0.31}
			&\makecell[c]{3.38}\\
			
			MSE Loss
			&\makecell[c]{10000}
			&\makecell[c]{76165}      
			&\makecell[c]{25.14}
			&\makecell[c]{29.33}
			&\makecell[c]{0.32}
			&\makecell[c]{3.40}\\
			
			\cellcolor[rgb]{1.0, 0.9490, 0.8} Huber Loss 
			&\cellcolor[rgb]{1.0, 0.9490, 0.8} \makecell[c]{10000}
			&\cellcolor[rgb]{1.0, 0.9490, 0.8} \makecell[c]{76165}        
			&\cellcolor[rgb]{1.0, 0.9490, 0.8} \makecell[c]{21.12}
			&\cellcolor[rgb]{1.0, 0.9490, 0.8} \makecell[c]{25.39}
			&\cellcolor[rgb]{1.0, 0.9490, 0.8} \makecell[c]{0.30}
			&\cellcolor[rgb]{1.0, 0.9490, 0.8} \makecell[c]{3.35}\\
			
			\Xhline{1pt}
			
		\end{tabular}
	}
	\vspace{-5pt} 
	\caption{The results on KITTI-360. Huber loss exhibits the better LiDAR generation quality for DDPMs targeting $\bm{v}$.}
	\label{supp_tab121}
	\vspace{-4mm}
\end{table}

\section{Additional Comparison Experiments}
\label{supp_sec2}

\subsection{Semantic-to-LiDAR Generation}

We also conduct Semantic-to-LiDAR generation on SemanticKITTI (23201 samples) \cite{behley2019semantickitti} (the implementation in Sec.~\ref{supp_sec5}). Tab.~\ref{supp_tab211} presents the results. By leveraging non-latent ControlNet \cite{zhang2023adding}, T2LDM demonstrates strong capability in generating LiDAR scenes conditioned on semantic maps. This further validates the effectiveness of ControlNet in non-latent DDPMs. Meanwhile, Fig.~\ref{supp_fig10} and Fig.~\ref{supp_fig11} further provides more visualization results for Semantic-to-LiDAR generation on SemanticKITTI \cite{behley2019semantickitti} and nuScenes \cite{caesar2020nuscenes}.

\vspace{-5pt}
\begin{table}[h]
\resizebox{0.475\textwidth}{!}{
		\begin{tabular}{p{3.0cm}|p{1.5cm}p{1.5cm}|p{1.1cm}p{1.1cm}p{1.2cm}p{1.1cm}}	
			\Xhline{1pt}
			
			{Methods}
			&\makecell[c]{Gen. Sam.}
			&\makecell[c]{Rea. Sam.}
			&\makecell[c]{FSVD$\downarrow$}
			&\makecell[c]{FPVD$\downarrow$}
			&\makecell[c]{JSD$\downarrow$}
			&\makecell[c]{MMD$\downarrow$}\\
			
			\Xhline{1pt}
			
			LiDM \cite{ran2024towards}
			&\makecell[c]{2000}
			&\makecell[c]{23201}      
			&\makecell[c]{201.41}
			&\makecell[c]{212.45}
			&\makecell[c]{0.33}
			&\makecell[c]{4.71}\\
			
			\cellcolor[rgb]{0.8352, 0.9098, 0.8314} T2LDM 
			&\cellcolor[rgb]{0.8352, 0.9098, 0.8314} \makecell[c]{2000}
			&\cellcolor[rgb]{0.8352, 0.9098, 0.8314} \makecell[c]{23201}        
			&\cellcolor[rgb]{0.8352, 0.9098, 0.8314} \makecell[c]{19.45}
			&\cellcolor[rgb]{0.8352, 0.9098, 0.8314} \makecell[c]{23.98}
			&\cellcolor[rgb]{0.8352, 0.9098, 0.8314} \makecell[c]{0.29}
			&\cellcolor[rgb]{0.8352, 0.9098, 0.8314} \makecell[c]{3.32}\\
			
			\Xhline{1pt}
			
		\end{tabular}
	}
	\vspace{-5pt} 
	\caption{The results on SemanticKITTI \cite{behley2019semantickitti}. T2LDM demonstrates excellent results for Semantic-to-LiDAR generation.}
	\label{supp_tab211}
	\vspace{-4mm}
\end{table}

\section{Conversion Derivations of \texorpdfstring{$\bm{v}$}{v}, \texorpdfstring{$\bm{x_0}$}{x_0}, and \texorpdfstring{$\bm{\epsilon}$}{\epsilon}}
\label{supp_sec3}

In this section, we logically derive the conversion relationships between $\bm{v}$, $\bm{x_0}$ and $\bm{\epsilon}$. Meanwhile, we also provide the explanation regarding the definition of $\bm{v}$.

\subsection{Motivation for Velocity Parameterization}

DDPMs typically predict $\bm{x_0}$ or $\bm{\epsilon}$. This uses the noise sample $\bm{x_t}=\bm{\mu_t}+\bm{\sigma_t}\bm{\epsilon}$ as input \cite{ho2020denoising, qu2024conditional, qu2025end}:

\vspace{-5pt}
\begin{equation}
	\begin{split}
		\label{supp_f311}
		\bm{x_t}=\sqrt{\overline{\alpha}_t}\bm{x_0} + \sqrt{1-\overline{\alpha}_t}\bm{\epsilon}.
	\end{split}
\end{equation}

However, the distributions of the target $\bm{x_0}$ and the target $\bm{\epsilon}$ remain constant in the training stage, \textit{i.e.}, $\bm{x_0} \sim \mathcal{P}_{data}$ and $\bm{\epsilon} \sim \mathcal{N}(\bm{0},\bm{I})$. \textit{\textbf{This means that the training process inevitably becomes unstable (the overly oscillatory loss), due to the varying SNR level of the input $\bm{x_t}$.}} 

\subsection{Definition of Velocity Parameterization}

To address the imbalance of SNR level between $\bm{x_0}$ prediction and $\bm{\epsilon}$ prediction caused by the timestep variation, a velocity-field target $\bm{v}$ combining $\bm{x_0}$ and $\bm{\epsilon}$ is introduced \cite{salimans2022progressive}. This is achieved by expressing $\bm{v}$ as a balanced combination relative to  $\bm{x_t}$, aiming to learn a smoother, more stable, and easier-to-optimize target:

\vspace{-5pt}
\begin{equation}
	\begin{split}
		\label{supp_f321}
		\bm{v}=\sqrt{\overline{\alpha}_t}\bm{\epsilon} - \sqrt{1-\overline{\alpha}_t}\bm{x_0}.
	\end{split}
\end{equation}

The velocity parameterization defines $\bm{v}$ as a complementary linear combination of $\bm{x_0}$ and $\bm{\epsilon}$ using the same coefficients as in the forward diffusion process. This construction ensures that the statistical structure of $\bm{v}$ aligns with the noise level of the input $\bm{x_t}$, providing a well-balanced target across timesteps. Moreover, this definition yields a invertible linear mapping between ($\bm{x_t}$, $\bm{v}$) and ($\bm{x_0}$, $\bm{\epsilon}$), enabling consistent denoising dynamics and improved training stability compared to $\bm{x_0}$ prediction and $\bm{\epsilon}$ prediction:

\vspace{-5pt}
\begin{equation}
	\begin{split}
		\label{supp_f322}
		\begin{pmatrix} x_t \\ v \end{pmatrix}
		=
		\begin{pmatrix} \sqrt{\bar{\alpha}_t} & \sqrt{1-\bar{\alpha}_t} \\ -\sqrt{1-\bar{\alpha}_t} & \sqrt{\bar{\alpha}_t} \end{pmatrix}
		\begin{pmatrix} x_0 \\ \epsilon \end{pmatrix},
	\end{split}
\end{equation}
where $\det\begin{pmatrix}\sqrt{\bar{\alpha}_t} & \sqrt{1-\bar{\alpha}_t} \\ -\sqrt{1-\bar{\alpha}_t} & \sqrt{\bar{\alpha}_t}\end{pmatrix}
= \bar{\alpha}_t + (1-\bar{\alpha}_t) = 1$. This means that ($\bm{x_t}$, $\bm{v}$) and ($\bm{x_0}$, $\bm{\epsilon}$) are invertible.

\subsection{Conversion between \texorpdfstring{$\bm{v}$}{v}, \texorpdfstring{$\bm{x_0}$}{x_0}, and \texorpdfstring{$\bm{\epsilon}$}{\epsilon}}

This conversion between $\bm{v}$, $\bm{x_0}$ and $\bm{\epsilon}$ mainly focuses on inference sampling. The original DDPM inference sampling is formulated as (the target $\bm{\epsilon}$) \cite{ho2020denoising, qu2025robust}:

\vspace{-5pt}
\begin{equation}
	\begin{split}
		\label{supp_f331}
		\bm{x_{t-1}}=\frac{1}{\sqrt{\alpha_t}}(\bm{x_t}-\frac{1-\alpha_t}{\sqrt{1-\overline{\alpha}_t}}\bm{\epsilon_{t-1}})\\
		+\sqrt{\frac{1-\overline{\alpha}_{t-1}}{1-\overline{\alpha}_t}(1-\alpha_t)} \bm{\epsilon}.\quad\;\;\;\;
	\end{split}
\end{equation}

Meanwhile, using Eq.~\ref{supp_f311}, Eq.~\ref{supp_f321} and Eq.~\ref{supp_f331}, $\bm{x_0}$ and $\bm{\epsilon}$ can be expressed in terms of $\bm{v}$ and $\bm{x_t}$:

\vspace{-5pt}
\begin{equation}
	\begin{split}
		\label{supp_f332}
		\bm{\epsilon} = \sqrt{1-\overline{a}_t}\bm{x_t} + \sqrt{\overline{a}_t}\bm{v}, \;\\
		\bm{x_0} = \sqrt{\overline{a}_t}\bm{x_t} - \sqrt{1-\overline{a}_t}\bm{v},\\
		\bm{x_0}=\frac{\bm{x_t}}{\sqrt{\overline{a}_t}}-\frac{\sqrt{1-\overline{a}_t}\bm{\epsilon}}{\sqrt{\overline{a}_t}}\;\;
	\end{split}
\end{equation}

Then, substituting Eq.~\ref{supp_f332} into Eq.~\ref{supp_f331} yields the inference sampling formulas for the target $\bm{x_0}$ and the target $\bm{v}$:

\vspace{-10pt}
\begin{equation}
	\begin{split}
		\label{supp_f333}
		\bm{x_{t-1}}=\frac{1}{\sqrt{\alpha_t}}\bm{x_0} + \sqrt{\frac{1-\overline{\alpha}_{t-1}}{1-\overline{\alpha}_t}(1-\alpha_t)} \bm{\epsilon},\quad\quad\\
		\bm{x_{t-1}} = \frac{1}{\sqrt{\alpha_t}} (\bm{x_t} - \frac{1-\alpha_t}{\sigma_t} [ \sigma_t \bm{x_t} + \sqrt{\bar{\alpha}_t} \, v_\theta] ) + \tilde{\sigma}_t \bm{\epsilon}
	\end{split}
\end{equation}
\vspace{-4pt}
\noindent where $\sigma_t=\sqrt{1-\bar{\alpha}_t}$, $\tilde{\sigma_t}=\sqrt{\frac{1 - \bar{\alpha}_{t-1}}{1 - \bar{\alpha}_t}(1-\alpha_t)}$. 

The above provides the conversion between $\bm{v}$, $\bm{x_0}$ and $\bm{\epsilon}$.

\begin{table*}[h]
	\resizebox{1.0\textwidth}{!}{
			\begin{tabular}{p{1.5cm}|p{2.5cm}|p{8.3cm}p{5.0cm}|p{8.3cm}}	
				\Xhline{1pt}
				
				{Level}
				&{Type}
				&{T2nuScenes Prompt Example}
				&{Sample Distribution}
				&{nuScenes Prompt Example}\\
				
				\Xhline{1pt}
				
				\multirow{10}{*}{($\color[rgb]{1.0,0.9020,0}{\blacksquare}$) Object}
				&\multirow{5}{*}{Quantity}
				&{($\color{blue}{\spadesuit}$) Two cars.}
				&{857,1873,2099,29320}
				&{Two parked motorcycles, overtake.}\\
				
				&
				&{($\color{blue}{\spadesuit}$) There are two cars in the scene.}
				&{857,1873,2099,29320}
				&{Go straight, two bicycles, turn right, two bendy busses.}\\
				
				&
				&{($\color{blue}{\spadesuit}$) Two cars. One car is in front. One car is behind.}
				&{857,1873,2099,29320}
				&{Two turning trucks, wait, turn left.}\\
				
				&
				&{($\color{red}{\clubsuit}$) Two cars. $\rightarrow$ There are two cars in the scene.}
				&{857,1873,2099,29320}
				&{two oncoming bikes, parked motorcycle, bike rack.}\\
				
				&
				&{($\color{red}{\clubsuit}$) There are two cars in the scene. $\rightarrow$ Two cars.}
				&{857,1873,2099,29320}
				&{Ped standing, ped sitting, two parked motorcycles on road.}\\
				
				&\cellcolor[rgb]{0.8549, 0.9098, 0.9882} Adjusted Text
				&\cellcolor[rgb]{0.8549, 0.9098, 0.9882} {($\color[rgb]{0.4392,0.1882,0.6275}{\blacklozenge}$) Less/More than five cars.}
				&\cellcolor[rgb]{0.8549, 0.9098, 0.9882} {10692,23457}
				&\cellcolor[rgb]{0.8549, 0.9098, 0.9882}{-}\\
				\cline{2-5}
				
				&{Location}
				&{One car is behind to the right of one pedestrian.}
				&{681,478,...,339,468 (41 texts)}
				&{Passing scooter, wait at intersection, turn right.}\\
				
				&\cellcolor[rgb]{1.0, 0.9020, 0.8} Adjusted Text
				&\cellcolor[rgb]{1.0, 0.9020, 0.8} {($\color[rgb]{0.4392,0.1882,0.6275}{\blacklozenge}$) No car./One car is around one pedestrian/barrier/truck.}
				&\cellcolor[rgb]{1.0, 0.9020, 0.8} 12227,11534,3819,6523
				&\cellcolor[rgb]{1.0, 0.9020, 0.8} -\\
				
				\cline{2-5}
				
				&{Orientation}
				&{One car is facing backward.}
				&{7416,6662,9994,8711,1366}
				&{-}\\
				
				&\cellcolor[rgb]{0.8352, 0.9098, 0.8314} Adjusted Text
				&\cellcolor[rgb]{0.8352, 0.9098, 0.8314} {($\color[rgb]{0.4392,0.1882,0.6275}{\blacklozenge}$) No car./One car is facing right/left.}
				&\cellcolor[rgb]{0.8352, 0.9098, 0.8314} {1366,16127,16656}
				&\cellcolor[rgb]{0.8352, 0.9098, 0.8314}{-}\\
				
				\hline
				
				\multirow{2}{*}{($\color[rgb]{0.9725,0.3216,0.8627}{\blacktriangle}$) Scene}
				&{Weather}
				&{Rainy/Sunny.}
				&{6670,27479}
				&{Night, rain, bump, peds, congestion, parked car.}\\
				
				\cline{2-5}
				
				&{Time}
				&{Night/Day.}
				&{3987,30162}
				&{Night, turn left, bumps, lightning.}\\
				
				\hline
				
				{-}
				&{Wea., Qua.}
				&{Rainy. Two cars.}
				&{1901,1706,...,205,42 (14 texts)}
				&{Rain, Wait at intersection, trucks, cars, ped.}\\
				
				{-}
				&\cellcolor[rgb]{1.0, 0.9490, 0.8} {Wea., Loc.}
				&\cellcolor[rgb]{1.0, 0.9490, 0.8} {($\color[rgb]{0.4392,0.6784,0.2784}{\bigstar}$) Rainy. One car is around one pedestrian.}
				&\cellcolor[rgb]{1.0, 0.9490, 0.8} {10101,9876,...,1658,853 (8 texts)}
				&\cellcolor[rgb]{1.0, 0.9490, 0.8}{Rain, turn left, turn left.}\\
				
				{-}
				&{Wea., Ori.}
				&{Rainy. One car is facing backward.}
				&{5740,5368,...,1294,114 (10 texts)}
				&{-}\\
				
				\cline{2-5}
				
				{-}
				&{Tim., Qua.}
				&{Night. Two cars.}
				&{21656,1962,...,302,206 (14 texts)}
				&{Night, ped in dark, parked cars.}\\
				
				{-}
				&{Tim., Loc.}
				&{Night. One car is around one pedestrian.}
				&{10140,10099,...,315,104 (8 texts)}
				&{Night, buses, peds, rain, right turn.}\\
				
				{-}
				&{Tim., Ori.}
				&{Night. One car is around one pedestrian.}
				&{8981,7655,...,695,464 (10 texts)}
				&{-}\\
				
				\cline{2-5}
				
				\multirow{2}{*}{-}
				&\multirow{2}{*}{Qua., Loc., Ori.}
				&{Two cars. One car is around one pedestrian.}
				&\multirow{2}{*}{2795,2152,...,24,7 (103 texts)}
				&\multirow{2}{*}{-}\\

				&
				&{One car is facing backward.}
				&
				&\\
				
				\cline{2-5}
				
				\multirow{2}{*}{-}
				&{Wea., Qua.,}
				&{Rainy, Two cars. One car is around one pedestrian.}
				&\multirow{2}{*}{1850,1182,...,2,1 (198 texts)}
				&\multirow{2}{*}{-}\\

				&{Loc., Ori.}
				&{One car is facing backward.}
				&
				&\\
				
				\cline{2-5}
				
				\multirow{2}{*}{-}
				&{Tim., Qua.,}
				&{Rainy, Two cars. One car is around one pedestrian.}
				&\multirow{2}{*}{2472,2013,...,1,1 (185 texts)}
				&\multirow{2}{*}{-}\\

				&{Loc., Ori.}
				&{One car is facing backward.}
				&
				&\\
				
				\hline
				
				\Xhline{1pt}
				
			\end{tabular}
		}
		\vspace{-5pt} 
		\caption{Results of different text forms. "Text1. $\rightarrow$ Text2." means that the model is trained with the text form of "Text2", while using "Text1" as conditional input in inference. "Wea., Loc." denotes "Weather, Location", exhibiting the more uniform sample distribution. Meanwhile, original text descriptions (the last column) in nuScenes are more unnatural than those (the third column) in T2nuScenes.} 
		\label{supp_tab421}
		\vspace{-1mm}
	\end{table*}

\section{Text-LiDAR Benchmark}
\label{supp_sec4}

In this section, we first describe the text annotation process of T2nuScenes and formalize the generation controllability metric (TBK). Subsequently, we provide the text-related sample distribution and the comparison between the original and re-annotated text descriptions in nuScenes \cite{caesar2020nuscenes}.

\textit{\textbf{Our primary goal is to provide a principled annotation strategy for LiDAR text descriptions, leveraging existing priors to construct high-quality Text–LiDAR pairs}}. We sincerely hope that researchers can further refine or develop improved text annotation methods in the future.

\subsection{Annotation Process}

In nuScenes \cite{caesar2020nuscenes}, we categorize the text descriptions into scene-level and object-level types. Meanwhile, \textit{\textbf{different types of text descriptions are stored separately, allowing them to be freely combined}}, as shown in Tab.~\ref{supp_tab421}. 

\textbf{For scene-level descriptions}, we reorganize the original text prompts from nuScenes according to the principles outlined in Sec.~3.3 of \textcolor{orange}{SM}. 

\textbf{\textcolor[rgb]{0.4392, 0.1882, 0.6275}{Weather.}} In the original nuScenes dataset, two weather conditions are provided: \textit{\textbf{Sunny}} and \textit{\textbf{Rainy}}.

\textbf{\textcolor[rgb]{0.9725,0.3216,0.8627}{Time.}} Similarly, the original nuScenes dataset divides time of a day into \textit{\textbf{Day}} and \textit{\textbf{Night}}.

\textbf{For object-level descriptions}, we leverage the 3D box priors and fix the target class as “car” to generate text descriptions that is closer to human natural language. This allows us to extract prior information about the quantity, location, and orientation of the target object in the scene.

\textbf{\textcolor[rgb]{0,0.6902,0.3137}{Quantity.}} The 3D boxes provide information about the number of the target object. Therefore, this can easily generate the following types of text descriptions regarding the quantity in the scene:

\begin{itemize}
	\item No target object.
	\item One/Two/... target object/s.
	\item There is/are one/two/... target object/s in the scene..
	\item One/Two/... target object/s. One target object is in front. Other target object/s is/are behind.
	\item \textcolor[rgb]{0,0.6902,0.3137}{Less/More than five target objects.}
	
\end{itemize}

As shown in Tab.~1 of \textcolor{orange}{MT}, overly detailed text prompts can significantly increase the training difficulty of the generative model. Therefore, we recommend using more coarse-grained text prompts.

\textbf{\textcolor[rgb]{1.0,0.7529,0}{Location.}} Meanwhile, since 3D boxes contain precise object coordinates, we can derive the target relative position to other objects in the scene. Given the target object box center ($cx_{tar}, cy_{tar}, cz_{tar}$) and another object box center ($cx_{ano}, cy_{ano}, cz_{ano}$), their relative position is defined:

\vspace{-5pt}
\begin{equation*}
	\begin{split}
		\label{supp_f411}
		(cx,cy,cz) = \quad\quad\quad\quad\quad\quad\quad\quad\quad\quad\quad\quad\quad\quad\quad\quad\quad\quad\\
		(cx_{tar}-cx_{ano}, cy_{tar}-cy_{ano}, cz_{tar}-cz_{ano}),\\
		p_1=
		\begin{cases}
			ahead,   & cx > threshold, \\
			behind,  & cx < -threhold, \\
			aligned, & -threhold \le cx \le threshold,
		\end{cases}\quad\quad\quad\\
	\end{split}
\end{equation*}

\vspace{-5pt}
\begin{equation}
	\begin{split}
		\label{supp_f411}
		p_2=
		\begin{cases}
			left,   & cy > threshold, \\
			right,  & cy < -threhold, \\
			center, & -threhold \le cy \le threshold,
		\end{cases}\quad\quad\quad\;\;\\
	\end{split}
\end{equation}
where the distance threshold is set 2.0m.

Subsequently, we can use $p_1$ and $p_2$ to form text descriptions of scene layouts, such as "One target object is $p_1$ to the $p_2$ of another object".

However, as mentioned in Sec.~3.3 of \textcolor{orange}{MT}, an overly dispersed sample distribution can significantly degrade generation quality and controllability. Therefore, instead of focusing on specific relative positions, \textit{\textbf{we only determine whether the target object and another object co-occur in the same scene}}.

Finally, this yields the following type of text description regarding the location in the scene:

\begin{itemize}
	\item No target object.
	\item One target object is ahead/behind/aligned to the left/right/center of another object.
	\item \textcolor[rgb]{1.0,0.7529,0}{One target object is around another object.}
	
\end{itemize}

\textbf{\textcolor{red}{Orientation.}} Furthermore, the 3D box also provides the orientation angle (in radians, $yaw$) of the target object. Therefore, we can also easily obtain the orientation of the target object in the scene: 

\vspace{-5pt}
\begin{equation}
	\begin{split}
		\label{supp_f412}
		deg = degrees(yaw), \quad\quad\quad\quad\quad\quad\quad\quad\quad\quad\quad\quad\;\\
		o=
		\begin{cases}
			forward,   & 315 \le deg \; or \; 45 > deg, \\
			left,      & 45 \le deg < 135, \\
			backward,  & 135 \le deg < 225, \\
			right,     & 225 \le deg < 315 \\
		\end{cases} \quad\quad\quad
	\end{split}
\end{equation}
where $degrees(\cdot)$ is a function that converts radians to degrees for consistent angular representation.

Then, we can use $o$ to describe the target object orientation in the scene, for example: "One target object is facing $o$". Therefore, we can generate orientation-based descriptions of the scene:
\begin{itemize}
	\item No target object.
	\item \textcolor{red}{One target object is facing forward/left/backward/right.}
	
\end{itemize}

The above process outlines how 3D box priors can be used to generate scene descriptions in natural language. Compared to directly using 3D boxes as conditions, natural language prompts are more accessible and flexible for providing semantic guidance to generative models.

\subsection{Generation Controllability Evaluation Metric}

We first utilize an existing 3D detector \cite{liu2025fshnet} to predict 3D boxes by detecting the generated LiDAR scenes. The matching \textbf{R}ate between the \textbf{T}ext prompts and the predicted 3D \textbf{B}oxes is used to measure the controllability of the generative model. For example, the text prompts for the three generated LiDAR scenes are: "Two cars", "One car", and "Five cars". The detector identified 1, 3, and 5 cars in the three LiDAR scenes, respectively. Thus, the \textbf{T}ext-to-\textbf{B}ox matching \textbf{R}ate (TBR) is 1/3$\approx$33.33$\%$.

Therefore, TBR can be formalized as: \textcolor{blue}{TBR} = \textcolor{red}{The Number of Correctly Matched LiDAR Scenes} $/$ \textcolor[rgb]{0,0.6902,0.3137}{The Number of Generated LiDAR Scenes}.

Benefiting from text descriptions derived from 3D box priors, we can quantitatively evaluate conditional controllability of generative models in Text-to-LiDAR generation.

\subsection{Sample Distributions}

In Tab.~\ref{supp_tab421}, we provide more detailed information regarding the distribution of text description samples compared to Tab.~1 in \textcolor{orange}{MT}. We can observe that the "Wea., Loc." combination exhibits a more uniform distribution compared to others, due to the minimum sample size of 853. This is significantly higher than other text combinations. Meanwhile, the diversity of the text descriptions is also crucial. Therefore, we consider "Wea., Loc." to be the optimal combination of text prompts in nuScenes.

\subsection{T2nuScenes vs. nuScenes}

Tab.~\ref{supp_tab421} also presents the comparison between the text descriptions of T2nuScenes and nuScenes. Clearly, the original text descriptions in nuScenes deviate significantly from natural human language. This hinders the Text-to-LiDAR generation effectiveness in training and the generalization of text conditions during inference, leading to the degradation in generation quality and controllability. Tab.~\ref{supp_tab441} further presents the comparative result of T2LDM generation performance on T2nuScenes and nuScenes. T2nuScenes demonstrates superior training effectiveness. 

\vspace{-7pt}
\begin{table}[h]
	\resizebox{0.475\textwidth}{!}{
		\begin{tabular}{p{3.0cm}|p{1.5cm}p{1.5cm}|p{1.1cm}p{1.1cm}p{1.2cm}p{1.1cm}}	
			\Xhline{1pt}
			
			{Methods}
			&\makecell[c]{Gen. Sam.}
			&\makecell[c]{Rea. Sam.}
			&\makecell[c]{FSVD$\downarrow$}
			&\makecell[c]{FPVD$\downarrow$}
			&\makecell[c]{JSD$\downarrow$}
			&\makecell[c]{MMD$\downarrow$}\\
			
			\Xhline{1pt}
			
			nuScenes \cite{caesar2020nuscenes}
			&\makecell[c]{10000}
			&\makecell[c]{34149}      
			&\makecell[c]{68.44}
			&\makecell[c]{66.92}
			&\makecell[c]{0.30}
			&\makecell[c]{3.05}\\
			
			\cellcolor[rgb]{0.8549, 0.9098, 0.9882} T2nuScenes 
			&\cellcolor[rgb]{0.8549, 0.9098, 0.9882} \makecell[c]{10000}
			&\cellcolor[rgb]{0.8549, 0.9098, 0.9882} \makecell[c]{34149}        
			&\cellcolor[rgb]{0.8549, 0.9098, 0.9882} \makecell[c]{66.93}
			&\cellcolor[rgb]{0.8549, 0.9098, 0.9882} \makecell[c]{65.84}
			&\cellcolor[rgb]{0.8549, 0.9098, 0.9882} \makecell[c]{0.28}
			&\cellcolor[rgb]{0.8549, 0.9098, 0.9882} \makecell[c]{3.05}
			\\
			
			\Xhline{1pt}
			
		\end{tabular}
	}
	\vspace{-3pt} 
	\caption{The text-guided results on nuScenes and T2nuScenes.}
	\label{supp_tab441}
	\vspace{-6mm}
\end{table}

\section{Implementation}
\label{supp_sec5}

In this section, we provide the hardware specifications and training time required for T2LDM. Meanwhile, we also describe the model hyperparameters  and the implementation of the non-latent ControlNet \cite{zhang2023adding} for T2LDM.

\subsection{Hardware requirement and Training Time}

We used \textit{\textbf{8 NVIDIA 4090 GPUs}} to train T2LDM on nuScenes \cite{caesar2020nuscenes}, KTIIT-360 \cite{liao2022kitti} and SemanticKITTI \cite{behley2019semantickitti}, which took approximately \textit{\textbf{47 hours, 58 hours and 57 hours}} on unconditional and conditional generation, respectively. For non-latent ControlNet \cite{zhang2023adding}, we fine-tuned T2LDM (only using frozen \textcolor{gray}{DN}) on nuScenes and SemanticKITTI about \textit{\textbf{35 hours and 48 hours}} for Sparse-to-Dense, Dense-to-Sparse, and Semantic-to-LiDAR generation.

\begin{table}[h]
	\scriptsize
	\resizebox{0.48\textwidth}{!}{
		\begin{tabular}{p{3.0cm}p{3.0cm}}	
			\Xhline{1pt}
			
			Config
			&\makecell[c]{Parameter} \\
			\cline{1-2}
			
			\textcolor[rgb]{0.9725,0.3216,0.8627}{T}
			&\makecell[c]{\textcolor[rgb]{0.9725,0.3216,0.8627}{1024}}\\
			
			\textcolor[rgb]{0.9725,0.3216,0.8627}{Time Embedding}
			&\makecell[c]{\textcolor[rgb]{0.9725,0.3216,0.8627}{Cos-Sin (384)}}\\
			
			\textcolor[rgb]{0.9725,0.3216,0.8627}{schedule}
			&\makecell[c]{\textcolor[rgb]{0.9725,0.3216,0.8627}{cosine}}\\
			
			\textcolor[rgb]{0.9725,0.3216,0.8627}{SNR gamma}
			&\makecell[c]{\textcolor[rgb]{0.9725,0.3216,0.8627}{5.0}}\\
			
			\textcolor[rgb]{0.9725,0.3216,0.8627}{diffusion target}
			&\makecell[c]{\textcolor[rgb]{0.9725,0.3216,0.8627}{$\bm{v}$}}\\
			
			\textcolor[rgb]{0.9725,0.3216,0.8627}{GN iterations}
			&\makecell[c]{\textcolor[rgb]{0.9725,0.3216,0.8627}{100k}}\\
			
			\textcolor[rgb]{0.9725,0.3216,0.8627}{$\lambda$}
			&\makecell[c]{\textcolor[rgb]{0.9725,0.3216,0.8627}{(0.001, 0.01, 0.1, 1.0)}}\\
			
			\textcolor[rgb]{0.9725,0.3216,0.8627}{$\lambda$ interval}
			&\makecell[c]{\textcolor[rgb]{0.9725,0.3216,0.8627}{25k}}\\
			
			\textcolor[rgb]{0.9725,0.3216,0.8627}{CFG scale}
			&\makecell[c]{\textcolor[rgb]{0.9725,0.3216,0.8627}{4.0}}\\
			
			\textcolor[rgb]{0.9725,0.3216,0.8627}{CFG dropout}
			&\makecell[c]{\textcolor[rgb]{0.9725,0.3216,0.8627}{0.1}}\\
			
			\hline
			
			\textcolor[rgb]{0,0.6902,0.3137}{CLIP model}
			&\makecell[c]{\textcolor[rgb]{0,0.6902,0.3137}{ViT-L/14}}\\
			
			\textcolor[rgb]{0,0.6902,0.3137}{CLIP channel}
			&\makecell[c]{\textcolor[rgb]{0,0.6902,0.3137}{768}}\\
			
			\hline
			
			\textcolor[rgb]{1.0,0.7529,0}{conv type}
			&\makecell[c]{\textcolor[rgb]{1.0,0.7529,0}{circular}}\\
			
			\textcolor[rgb]{1.0,0.7529,0}{base channel}
			&\makecell[c]{\textcolor[rgb]{1.0,0.7529,0}{64}}\\
			
			\textcolor[rgb]{1.0,0.7529,0}{channels}
			&\makecell[c]{\textcolor[rgb]{1.0,0.7529,0}{(1,2,4,4)}}\\
			
			\textcolor[rgb]{1.0,0.7529,0}{strides}
			&\makecell[c]{\textcolor[rgb]{1.0,0.7529,0}{(1, 2), (2, 2), (2, 2)}}\\
			
			\textcolor[rgb]{1.0,0.7529,0}{attention types}
			&\makecell[c]{\textcolor[rgb]{1.0,0.7529,0}{(linear, linear, linear, vanilla)}}\\
			
			\textcolor[rgb]{1.0,0.7529,0}{attention heads}
			&\makecell[c]{\textcolor[rgb]{1.0,0.7529,0}{(2,4,8,8)}}\\
			
			\textcolor[rgb]{1.0,0.7529,0}{attention pe}
			&\makecell[c]{\textcolor[rgb]{1.0,0.7529,0}{rope}}\\
			
			\textcolor[rgb]{1.0,0.7529,0}{skip connection scale}
			&\makecell[c]{\textcolor[rgb]{1.0,0.7529,0}{sqrt(2)}}\\
			
			\textcolor[rgb]{1.0,0.7529,0}{norm types}
			&\makecell[c]{\textcolor[rgb]{1.0,0.7529,0}{(group, group, group, group)}}\\
			
			\textcolor[rgb]{1.0,0.7529,0}{position encoding}
			&\makecell[c]{\textcolor[rgb]{1.0,0.7529,0}{dpe}}\\
			
			\hline
			
			\textcolor{red}{use ema}
			&\makecell[c]{\textcolor{red}{True}}\\
			
			\textcolor{red}{ema decay}
			&\makecell[c]{\textcolor{red}{0.9997}}\\
			
			\textcolor{red}{ema update}
			&\makecell[c]{\textcolor{red}{1}}\\
			
			\Xhline{1pt}
			
		\end{tabular}
	}
	\vspace{-3pt} 
	\caption{The parameters of network framework for T2LDM.}
	\label{supp_tab521}
	\vspace{-4mm}
\end{table}

\begin{table}[h]
	\scriptsize
	\resizebox{0.48\textwidth}{!}{
		\begin{tabular}{p{1.5cm}p{1.2cm}p{0.0005cm}p{1.5cm}p{1.2cm}p{0.0005cm}p{1.5cm}p{1.2cm}}	
			\Xhline{1pt}
			
			\multicolumn{2}{c}{{\textcolor[rgb]{0,0.6902,0.3137}{nuScenes}} \cite{caesar2020nuscenes}}
			&\quad
			&\multicolumn{2}{c}{{\textcolor[rgb]{1.0,0.7529,0}{KITTI-360}} \cite{liao2022kitti}}
			&\quad
			&\multicolumn{2}{c}{\textcolor{red}{SemanticKITTI} \cite{behley2019semantickitti}}
			\\
			\cline{1-2} \cline{4-5} \cline{7-8} 
			
			{\textcolor[rgb]{0,0.6902,0.3137}{Config}}
			&\makecell[c]{\textcolor[rgb]{0,0.6902,0.3137}{Parameter}} 
			&\quad
			&{\textcolor[rgb]{1.0,0.7529,0}{Config}}
			&\makecell[c]{\textcolor[rgb]{1.0,0.7529,0}{Parameter}}
			&\quad
			&\textcolor{red}{Config}
			&\makecell[c]{\textcolor{red}{Parameter}}
			\\
			\hline
			
			\textcolor[rgb]{0,0.6902,0.3137}{Optimizer}
			&\makecell[c]{\textcolor[rgb]{0,0.6902,0.3137}{Adam}} 
			&\quad
			&\textcolor[rgb]{1.0,0.7529,0}{Optimizer}
			&\makecell[c]{\textcolor[rgb]{1.0,0.7529,0}{Adam}}
			&\quad
			&\textcolor{red}{Optimizer}
			&\makecell[c]{\textcolor{red}{Adam}}
			\\
			
			\textcolor[rgb]{0,0.6902,0.3137}{Scheduler}
			&\makecell[c]{\textcolor[rgb]{0,0.6902,0.3137}{LR Cosine}} 
			&\quad
			&\textcolor[rgb]{1.0,0.7529,0}{Scheduler}
			&\makecell[c]{\textcolor[rgb]{1.0,0.7529,0}{LR Cosine}} 
			&\quad
			&\textcolor{red}{Scheduler}
			&\makecell[c]{\textcolor{red}{LR Cosine}} 
			\\
			
			\textcolor[rgb]{0,0.6902,0.3137}{LR}
			&\makecell[c]{\textcolor[rgb]{0,0.6902,0.3137}{1e-4}} 
			&\quad
			&\textcolor[rgb]{1.0,0.7529,0}{LR}
			&\makecell[c]{\textcolor[rgb]{1.0,0.7529,0}{1e-4}}
			&\quad
			&\textcolor{red}{LR}
			&\makecell[c]{\textcolor{red}{1e-4}}
			\\
			
			\textcolor[rgb]{0,0.6902,0.3137}{Weight De.}
			&\makecell[c]{\textcolor[rgb]{0,0.6902,0.3137}{0.01}} 
			&\quad
			&\textcolor[rgb]{1.0,0.7529,0}{Weight De.}
			&\makecell[c]{\textcolor[rgb]{1.0,0.7529,0}{0.01}}
			&\quad
			&\textcolor{red}{Weight De.}
			&\makecell[c]{\textcolor{red}{0.01}}
			\\
			
			\textcolor[rgb]{0,0.6902,0.3137}{Batch Size}
			&\makecell[c]{\textcolor[rgb]{0,0.6902,0.3137}{16}} 
			&\quad
			&\textcolor[rgb]{1.0,0.7529,0}{Batch Size}
			&\makecell[c]{\textcolor[rgb]{1.0,0.7529,0}{16}}
			&\quad
			&\textcolor{red}{Batch Size}
			&\makecell[c]{\textcolor{red}{16}}
			\\
			
			\textcolor[rgb]{0,0.6902,0.3137}{Iterations}
			&\makecell[c]{\textcolor[rgb]{0,0.6902,0.3137}{400K}} 
			&\quad
			&\textcolor[rgb]{1.0,0.7529,0}{Iterations}
			&\makecell[c]{\textcolor[rgb]{1.0,0.7529,0}{400K}}
			&\quad
			&\textcolor{red}{Iterations}
			&\makecell[c]{\textcolor{red}{400K}}
			\\
			\hline
			
			\textcolor[rgb]{0,0.6902,0.3137}{resolution}
			&\makecell[c]{\textcolor[rgb]{0,0.6902,0.3137}{(32,1024)}} 
			&\quad
			&\textcolor[rgb]{1.0,0.7529,0}{resolution}
			&\makecell[c]{\textcolor[rgb]{1.0,0.7529,0}{(64,1024)}}
			&\quad
			&\textcolor{red}{resolution}
			&\makecell[c]{\textcolor{red}{(64,1024)}}
			\\
			
			\textcolor[rgb]{0,0.6902,0.3137}{depth range}
			&\makecell[c]{\textcolor[rgb]{0,0.6902,0.3137}{(0.01,50.0)}} 
			&\quad
			&\textcolor[rgb]{1.0,0.7529,0}{depth range}
			&\makecell[c]{\textcolor[rgb]{1.0,0.7529,0}{(1.45,80.0)}}
			&\quad
			&\textcolor{red}{depth range}
			&\makecell[c]{\textcolor{red}{(1.45,80.0)}}
			\\
			
			\textcolor[rgb]{0,0.6902,0.3137}{fov}
			&\makecell[c]{\textcolor[rgb]{0,0.6902,0.3137}{(-3,25)}} 
			&\quad
			&\textcolor[rgb]{1.0,0.7529,0}{fov}
			&\makecell[c]{\textcolor[rgb]{1.0,0.7529,0}{(-3,25)}} 
			&\quad
			&\textcolor{red}{fov}
			&\makecell[c]{\textcolor{red}{(-3,25)}} 
			\\
			
			\textcolor[rgb]{0,0.6902,0.3137}{use intensity}
			&\makecell[c]{\textcolor[rgb]{0,0.6902,0.3137}{True}} 
			&\quad
			&\textcolor[rgb]{1.0,0.7529,0}{use intensity}
			&\makecell[c]{\textcolor[rgb]{1.0,0.7529,0}{True}} 
			&\quad
			&\textcolor{red}{use intensity}
			&\makecell[c]{\textcolor{red}{True}}  
			\\
			
			\Xhline{1pt}
			
		\end{tabular}
	}
	
	\vspace{-3pt} 
	\caption{The training hyperparameters of T2LDM.}
	\label{supp_tab522}
	\vspace{-4mm}
\end{table}

\subsection{Model Hyperparameters}

We implement a highly flexible network framework that constructs a corresponding U-Net architecture from input lists of Encoder, Middle, and Decoder, as illustrated in Fig.~\ref{supp_fig1}(left). T2LDM adopts the U-Net architecture from Stable Diffusion \cite{rombach2022high}. Meanwhile, due to the limited sample size of LiDAR datasets, an excessive number of parameters can easily lead to model collapse, resulting in over-smoothed and homogeneous generation results, as seen in LiDM \cite{ran2024towards}. Therefore, T2lDM employs a smaller channel dimension than Stable Diffusion, as shown in Fig.~\ref{supp_fig1}(left).

Furthermore, the detailed parameters of the network architecture of T2LDM are described in Tab.~\ref{supp_tab521}, while the training hyperparameters for each benchmark (nuScenes, KITTI-360, SemanticKITTI) are shown in Tab.~\ref{supp_tab522}.

\subsection{Network Architecture}

\begin{figure*}[htp]
	\centering
	\includegraphics[width=0.99\textwidth]{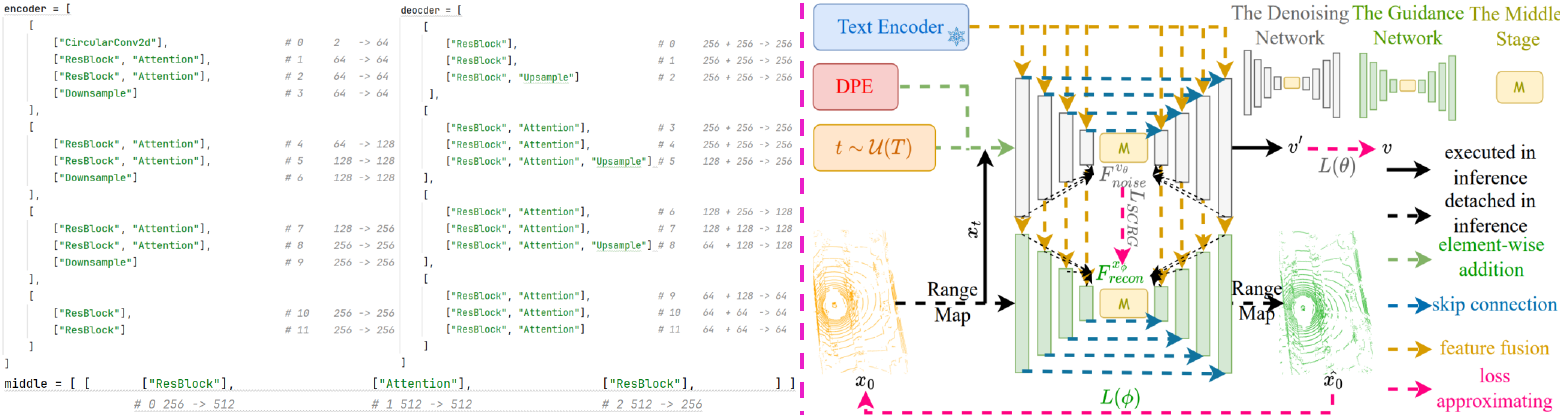}
	\vspace{-0.3cm}
	\caption{The overall framework of T2LDM. \textcolor{gray}{DN} and \textcolor[rgb]{0,0.6902,0.3137}{GN} are composed of an Encoder, a Middle Stage, and a Decoder. The core modules of T2LDM include: ResBlock (RB), AttentionBlock (AB), DownsamplingBlock (DB), and UpsamplingBlock (UB). To effectively process the spherical projection of LiDAR data, T2LDM incorporates Circular Convolution \cite{stearns2024curvecloudnet} to adapt to the unfolded Range Map.}
	\label{supp_fig1}
	
\end{figure*}

In Fig.~\ref{supp_fig1}, Encoder, Middle, and Decoder consist of four modules: \textit{\textbf{ResBlock (RB),  AttentionBlock (AB), DownsamplingBlock (DB), and UpsamplingBlock (UB)}}.

\textbf{ResBlock.} RB consists of two GroupNorm layers, two activation layers, two circular convolution \cite{stearns2024curvecloudnet} with a residual skip. We adopt the Swish activation $f(x)=x \cdot sigmoid(x)$. Meanwhile, the time embedding is injected through a MLP layers. The output is computed as $x + h$. 

\vspace{-5pt}
\begin{equation}
	\begin{split}
		\label{supp_f531}
		h = Conv_1(Swish_1(GN_1(x))),\\
		h = h + mlp(temb),\quad\quad\;\;\\
		h = Conv_2(Swish_2(GN_2(h))),\\
		x = h + x.\quad\quad\quad\quad\;\;\\
	\end{split}
\end{equation}

Following Stable Diffusion \cite{rombach2022high}, we apply zero-initialization to the $Conv_2(\cdot)$ layers used the skip connections to enhance training stability.

\textbf{AttentionBlock.} T2LDM employs linear AB and conventional AB for unconditional and conditional generation.

For linear AB (unconditional generation), given \textcolor[rgb]{0,0.6902,0.3137}{$F^{v_{\theta}}_{noise} \in \mathbb{R}^{l \times C^{v_{\theta}}}$$\rightarrow$$(Q,K,V) \in \mathbb{R}^{C \times l}$} via mlps, a self-attention block is conducted:

\vspace{-8pt}
\begin{equation}
	\begin{split}
		\label{supp_f531}
		O=(\frac{softmax(K)V^{T}}{\sqrt{C}})Q+F^{v_{\theta}}_{noise},\\
		F=ffn(O)+O. \quad\quad\quad\quad
	\end{split}
\end{equation}

For conventional AB (conditional generation), given \textcolor{red}{$F^{v_{\theta}}_{noise} \in \mathbb{R}^{l \times C^{v_{\theta}}}$$\rightarrow$$(Q) \in \mathbb{R}^{l \times C}$} and \textcolor{red}{$F^{CLIP}_{text} \in \mathbb{R}^{n \times 768}$$\rightarrow$$(K,V) \in \mathbb{R}^{n \times C}$} via mlps, a cross-attention block is conducted:

\vspace{-12pt}
\begin{equation}
	\begin{split}
		\label{supp_f532}
		O=softmax(\frac{QK^{T}}{\sqrt{C}})V+F^{v_{\theta}}_{noise},\\
		F=ffn(O)+O. \quad\quad\quad\quad
	\end{split}
\end{equation}

\begin{figure}[htp]
	\centering
	\includegraphics[width=0.475\textwidth]{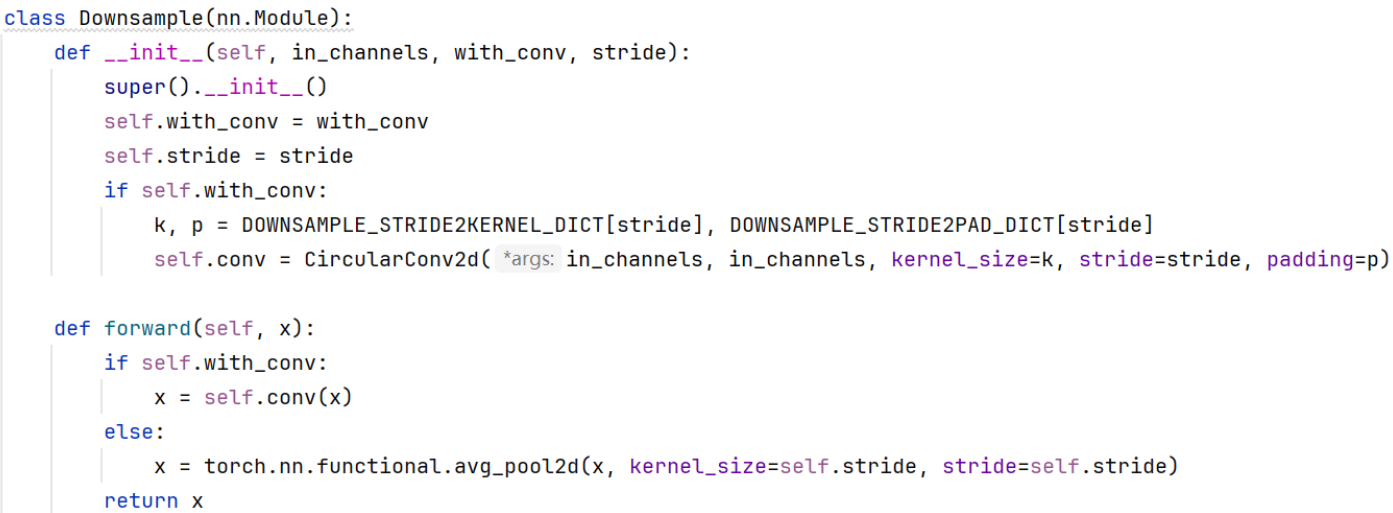}
	\vspace{-0.6cm}
	\caption{The code of the downsampling Block.}
	\label{supp_fig2}
	
\end{figure}

\begin{figure}[htp]
	\centering
	\includegraphics[width=0.475\textwidth]{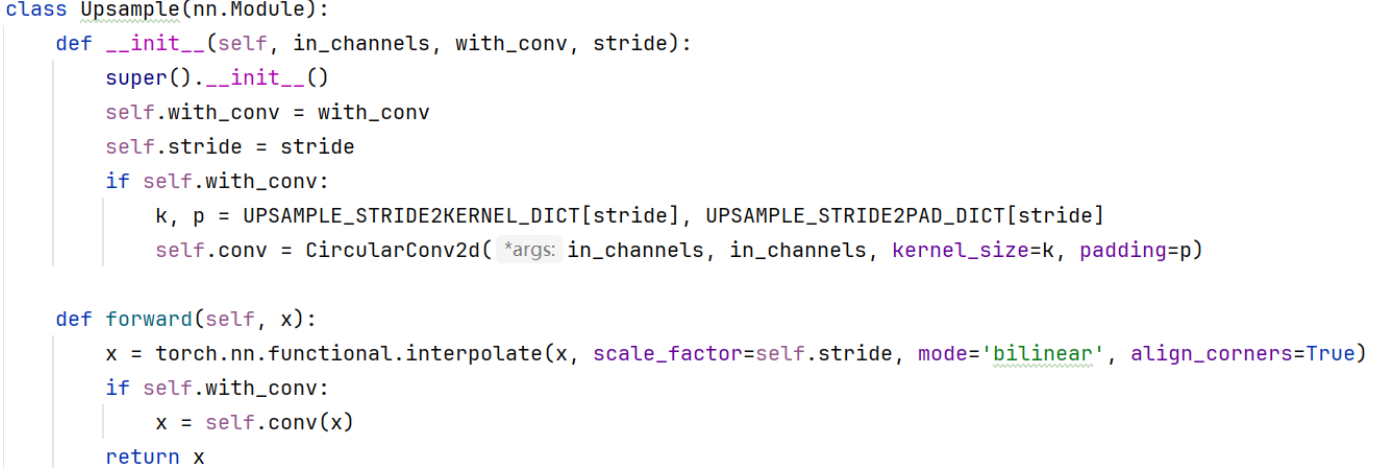}
	\vspace{-0.6cm}
	\caption{The code of the upsampling Block.}
	\label{supp_fig3}
	
\end{figure}

\begin{figure}[htp]
	\centering
	\includegraphics[width=0.475\textwidth]{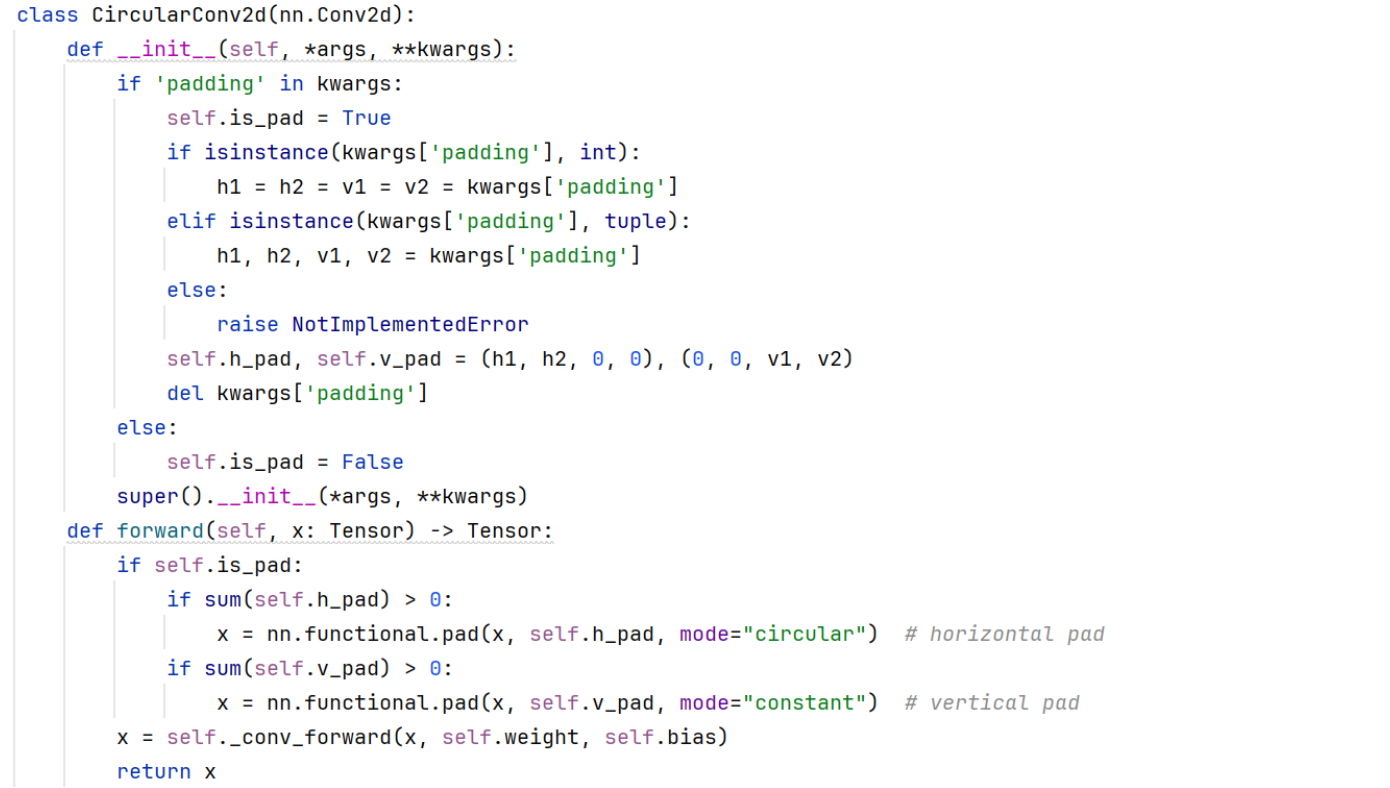}
	\vspace{-0.6cm}
	\caption{The code of Circular Convolution.}
	\label{supp_fig4}
	\vspace{-4mm}
\end{figure}

$l=h \times w$. Linear AB operates attention along the channel dimension (\textcolor[rgb]{0,0.6902,0.3137}{$W \in \mathbb{R}^{C \times l}$}), resulting in linear computational complexity. In contrast,  conventional AB integrates similar features in the spatial dimension (\textcolor{red}{$W \in \mathbb{R}^{l \times C}$}), enhancing contextual modeling capacity.

\textbf{DownsamplingBlock.}  DB progressively downsamples the range map using 2D pooling. The corresponding code is provided in Fig.~\ref{supp_fig2}.

\textbf{UpsamplingBlock.}  UB progressively upsamples the range map using bilinear interpolation. The corresponding code is provided in Fig.~\ref{supp_fig3}.

\begin{figure}[htp]
	\centering
	\includegraphics[width=0.475\textwidth]{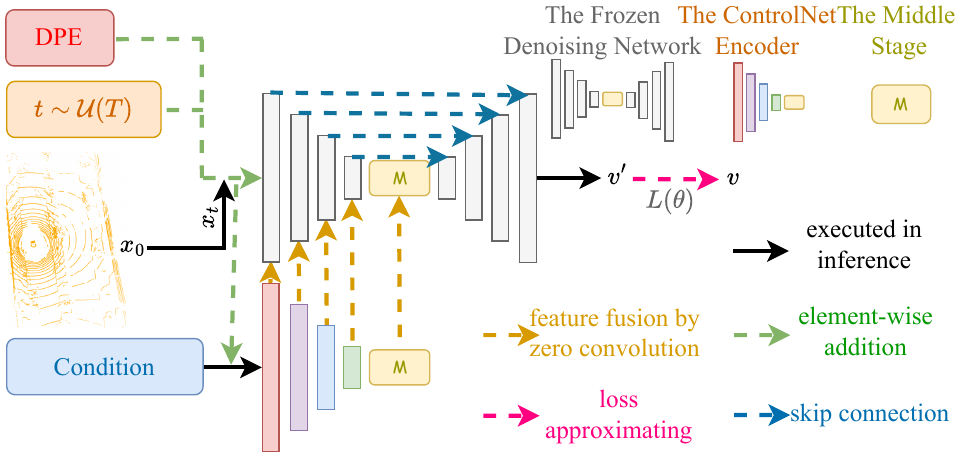}
	\vspace{-0.6cm}
	\caption{The overall framework of non-latent ControlNet.}
	\label{supp_fig5}
	\vspace{-5mm}
\end{figure}

\textbf{Circular Convolution.} Circular Convolution applies circular padding along the horizontal axis and zero padding vertically to preserve the periodic structure of range-view inputs. Compare to traditional convolution, this eliminates boundary discontinuities at the $0^\circ/360^\circ$ transition. The corresponding code is provided in Fig.~\ref{supp_fig4}.

\subsection{Non-Latent ControlNet}

To the best of our knowledge, \textit{\textbf{we present the first attempt to integrate ControlNet \cite{zhang2023adding} into non-latent DDPMs in 3D generation}}. Similar to the approach in latent ControlNet \cite{zhang2023adding}, we freeze \textcolor{gray}{DN} of unconditional T2LDM and introduce an additional encoder (\textbf{\textcolor[rgb]{0.9725,0.3216,0.8627}{$\mathcal{E}_c$}}), which shares the same architecture as the encoder in \textcolor{gray}{DN}, to learn the conditional features. Experimental results demonstrate that T2LDM, leveraging the non-latent ControlNet,  can achieves significant controllable generation (see Sec.~5.4 of \textcolor{orange}{MT}, Fig.~\ref{supp_fig10}, and Fig.~\ref{supp_fig11}).

\textbf{Sparse-to-Dense/Dense-to-Sparse Generation.} During training, \textbf{\textcolor[rgb]{0.9725,0.3216,0.8627}{$\mathcal{E}_c$}} takes sparse point clouds as input to extract conditional features (the process of sparse point clouds in Sec.~5.4 of \textcolor{orange}{MT}). Subsequently, zero convolution layers are utilized to aggregate features from corresponding layers in \textcolor{gray}{DN} and \textbf{\textcolor[rgb]{0.9725,0.3216,0.8627}{$\mathcal{E}_c$}}. The skip connections inject conditional information from \textbf{\textcolor[rgb]{0.9725,0.3216,0.8627}{$\mathcal{E}_c$}} into the decoder of \textcolor{gray}{DN}. Finally, the reconstruction loss from \textcolor{gray}{DN} is back-propagated to \textbf{\textcolor[rgb]{0.9725,0.3216,0.8627}{$\mathcal{E}_c$}} to update the parameters. Fig.~\ref{supp_fig5} illustrates the overall framework. Meanwhile, since the output LiDAR data shape is determined by the input noise size, T2LDM can perform upsampling or downsampling at arbitrary rates, as demonstrated in Fig.~6 of \textcolor{orange}{MT}. Furthermore, following the inference track of PUDM \cite{qu2024conditional}, T2LDM further improves generation results.

\textbf{Semantic-to-LiDAR Generation.} Similarly, following an analogous procedure, we feed the normalized semantic map (Semantic Map / Class Num) into \textbf{\textcolor[rgb]{0.9725,0.3216,0.8627}{$\mathcal{E}_c$}} to achieve LiDAR scene generation conditioned on the semantic map.

\vspace{-5pt}
\section{Limitations and Future Works}
\label{supp_sec6}

\subsection{Limited Generality for Text Annotation}

Although leveraging object detection priors to generate LiDAR scene text descriptions is effective, most existing LiDAR datasets (e.g., KITTI-360 \cite{liao2022kitti}) lack 3D box annotations. This inspires us to develop alternative priors for generating effective scene descriptions.

\subsection{Lack of Multi-Conditional Control}

Currently, T2LDM can only achieve effective controllable generation with a single condition. However, existing LiDAR data often possesses multiple annotation priors, such as semantic maps and 3D bounding boxes. In future work, we will integrate multi-conditional controllable generation into T2LDM.

\subsection{Increased Parameter Demand}

Despite SCRG only requiring a \textcolor[rgb]{0,0.6902,0.3137}{Guidance Network (GN)}, sharing the same architecture as the \textcolor{gray}{Denoising Network (DN)}, for joint training in the early stages and being decoupled during inference, the doubled parameter count still imposes high resource demands. Therefore, the future research will explore the path of pruning \textcolor[rgb]{0,0.6902,0.3137}{GN} to a lightweight version, alleviating computational overhead.

\section{More  Visualization Results}
\label{supp_sec7}

\begin{figure*}[htp]
	\centering
	\includegraphics[width=0.95\textwidth]{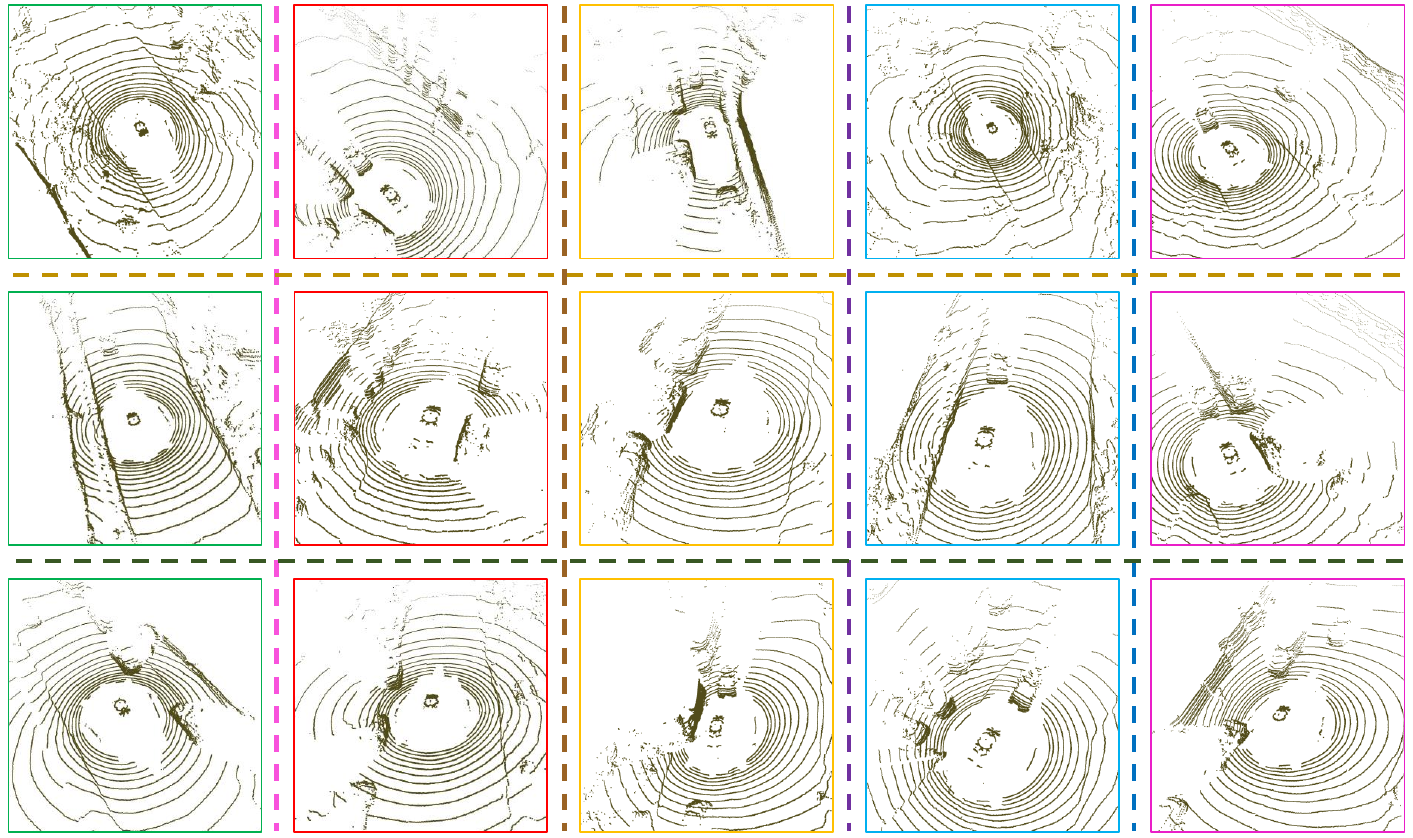}
	\vspace{-0.1cm}
	\caption{The visualization of unconditional generation for T2LDM on nuScenes \cite{caesar2020nuscenes}. }
	\label{supp_fig6}
	
\end{figure*}

\begin{figure*}[htp]
	\centering
	\includegraphics[width=0.95\textwidth]{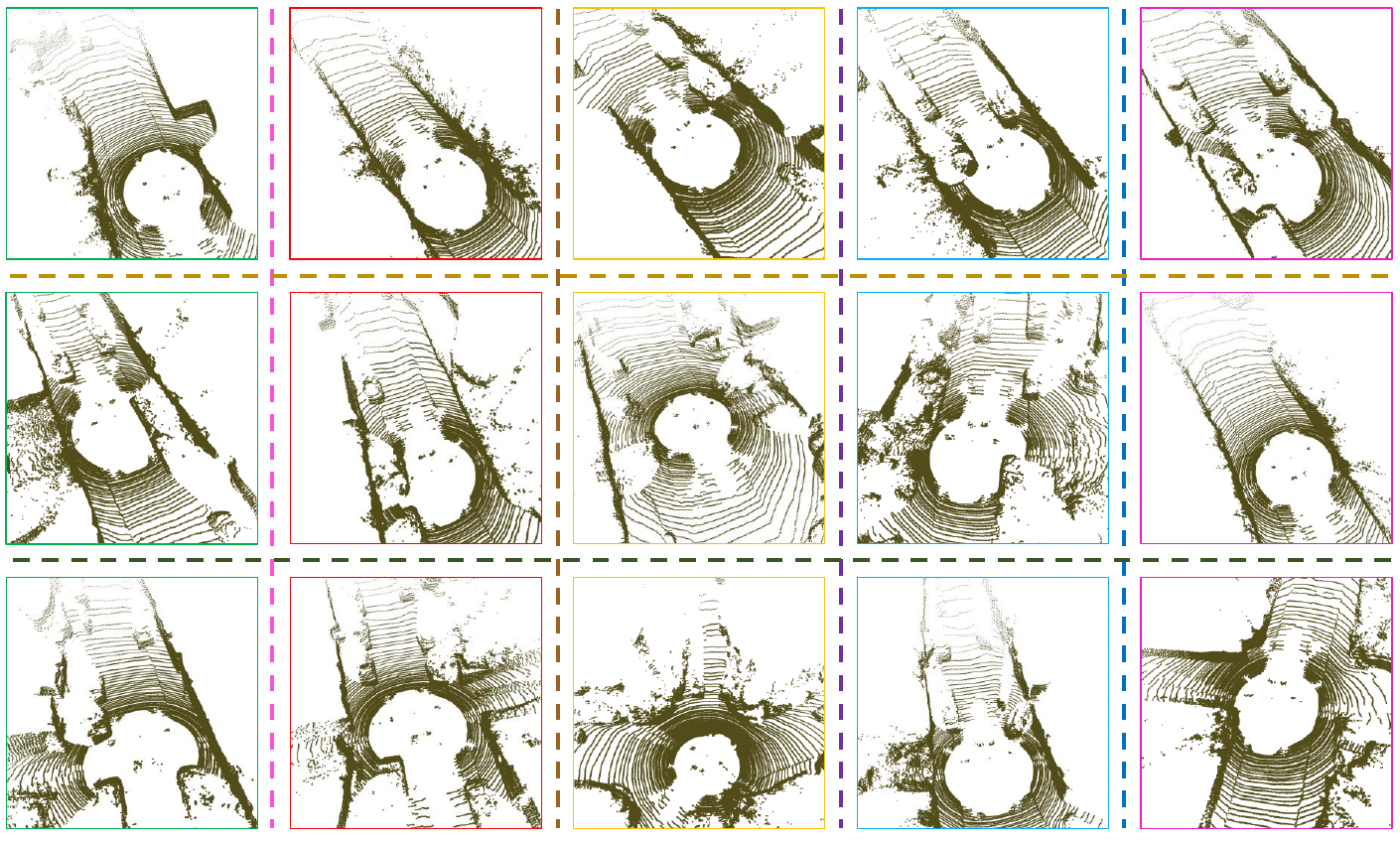}
	\vspace{-0.1cm}
	\caption{The visualization of unconditional generation for T2LDM on KITTI-360 \cite{liao2022kitti}. }
	\label{supp_fig7}
\end{figure*}

\begin{figure*}[htp]
	\centering
	\includegraphics[width=0.95\textwidth]{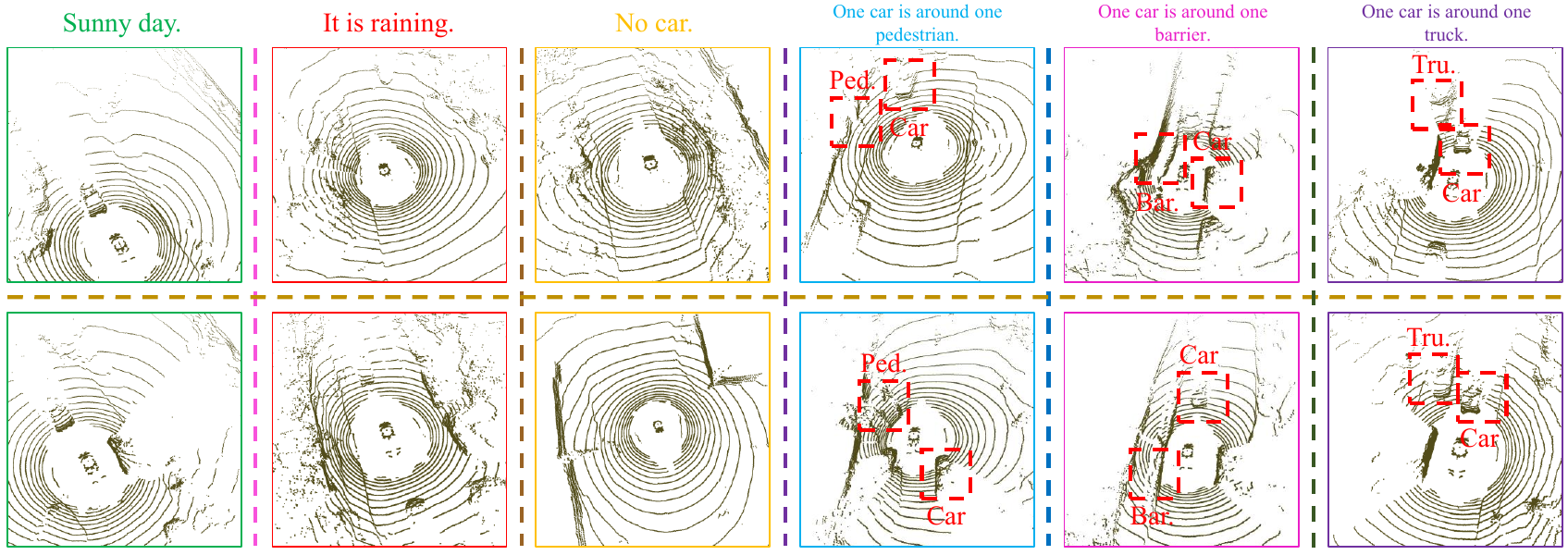}
	\vspace{-0.1cm}
	\caption{The visualization of text-guided generation for T2LDM on nuScenes \cite{caesar2020nuscenes}. }
	\label{supp_fig8}
\end{figure*}

\begin{figure*}[htp]
	\centering
	\includegraphics[width=0.95\textwidth]{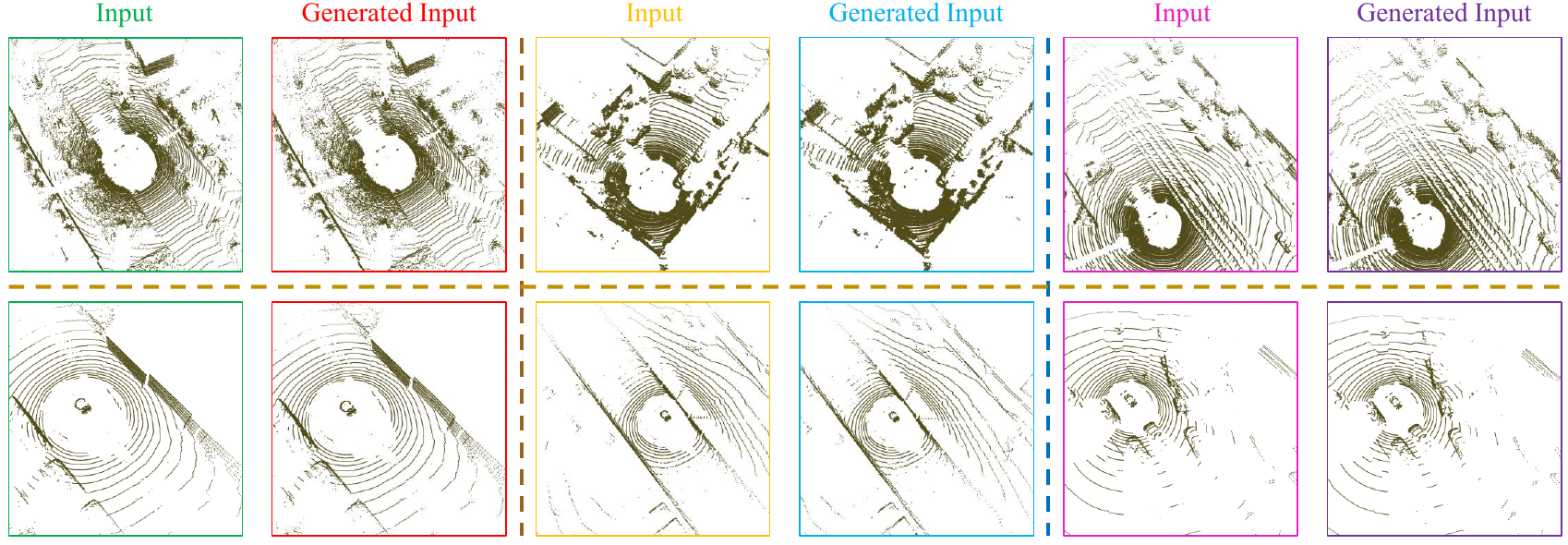}
	\vspace{-0.1cm}
	\caption{The visualization of reconstruction input via \textcolor[rgb]{0,0.6902,0.3137}{DN} on KITTI-360 \cite{liao2022kitti} (up) and nuScenes \cite{caesar2020nuscenes} (bottom). }
	\label{supp_fig9}
\end{figure*}

\begin{figure*}[htp]
	\centering
	\includegraphics[width=0.95\textwidth]{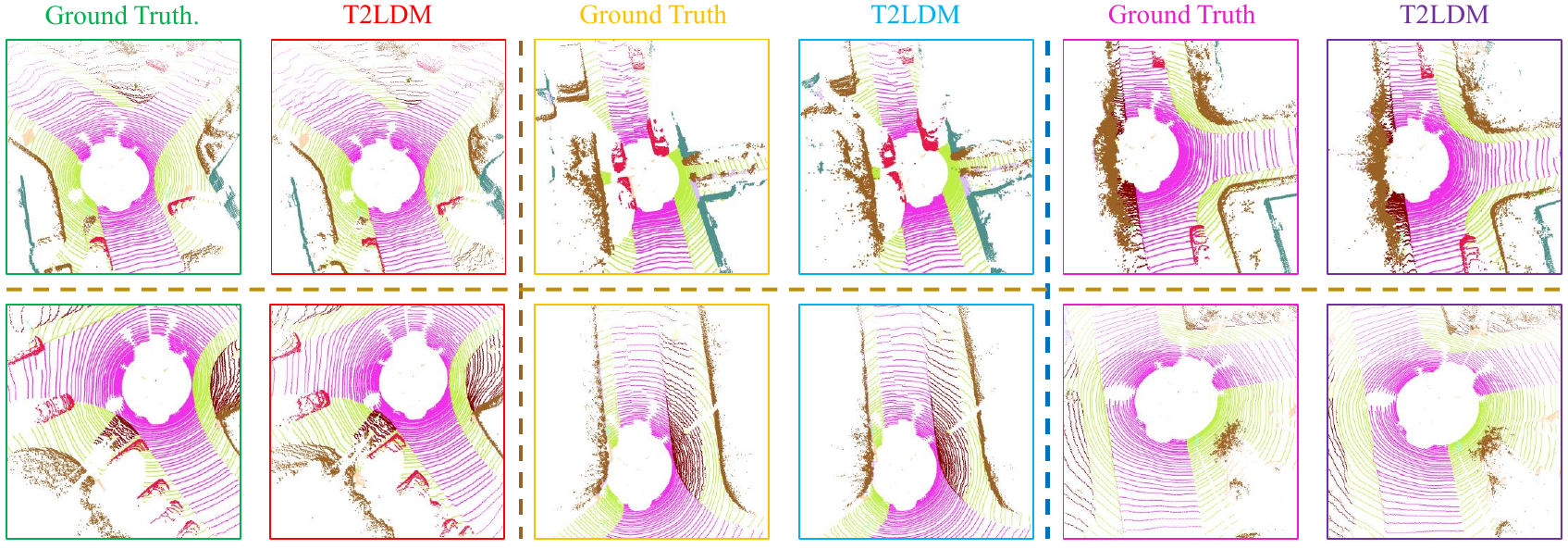}
	\vspace{-0.1cm}
	\caption{The visualization of Semantic-to-LiDAR generation for T2LDM on SemanticKITTI \cite{behley2019semantickitti}. }
	\label{supp_fig10}
\end{figure*}

\begin{figure*}[htp]
	\centering
	\includegraphics[width=0.95\textwidth]{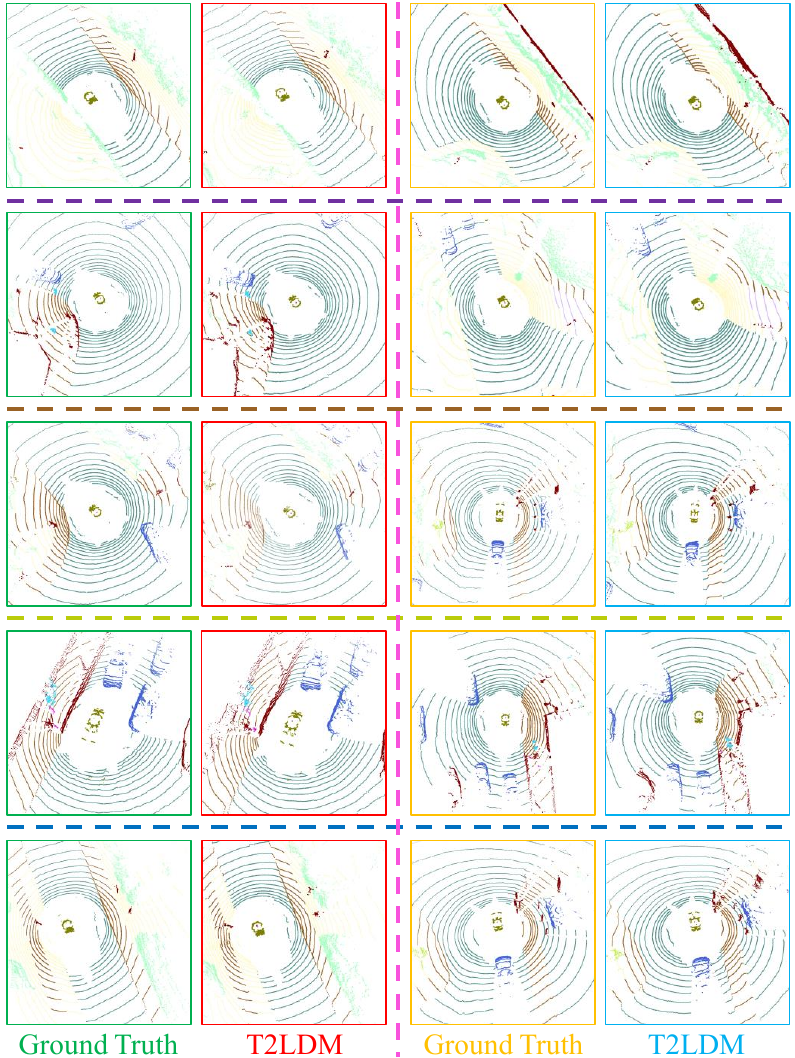}
	\vspace{-0.1cm}
	\caption{The visualization of Semantic-to-LiDAR generation for T2LDM on nuScenes \cite{caesar2020nuscenes}. }
	\label{supp_fig11}
\end{figure*}

We provide additional generation visualization results:

\begin{itemize}
	\item Fig.~\ref{supp_fig6} (unconditional generation on nuScenes \cite{caesar2020nuscenes})
	
	\item Fig.~\ref{supp_fig7} (unconditional generation on KITTI-360 \cite{liao2022kitti})
	
	\item Fig.~\ref{supp_fig8} (text-guided generation on nuScenes \cite{caesar2020nuscenes})
	
	\item Fig.~\ref{supp_fig9} (reconstruction input via \textcolor[rgb]{0,0.6902,0.3137}{DN}) on KITTI-360 \cite{liao2022kitti} (up) and nuScenes \cite{caesar2020nuscenes} (bottom)
	
	\item Fig.~\ref{supp_fig10} (Semantic-to-LiDAR generation on SemanticKITTI \cite{behley2019semantickitti})
	
	\item Fig.~\ref{supp_fig11} (Semantic-to-LiDAR generation on nuScenes \cite{caesar2020nuscenes})
	
\end{itemize}

\newpage

\newpage

\clearpage

{
	\small
	\bibliographystyle{ieeenat_fullname}
	\bibliography{main}

@article{li2020deep,
	title={Deep learning for lidar point clouds in autonomous driving: A review},
	author={Li, Ying and Ma, Lingfei and Zhong, Zilong and Liu, Fei and Chapman, Michael A and Cao, Dongpu and Li, Jonathan},
	journal={IEEE Transactions on Neural Networks and Learning Systems},
	volume={32},
	number={8},
	pages={3412--3432},
	year={2020},
	publisher={IEEE}
}

@inproceedings{behari2025blurred,
	title={Blurred lidar for sharper 3d: Robust handheld 3d scanning with diffuse lidar and rgb},
	author={Behari, Nikhil and Young, Aaron and Somasundaram, Siddharth and Klinghoffer, Tzofi and Dave, Akshat and Raskar, Ramesh},
	booktitle={Proceedings of the Computer Vision and Pattern Recognition Conference},
	pages={26954--26964},
	year={2025}
}

@article{yin2024survey,
	title={A survey on global lidar localization: Challenges, advances and open problems},
	author={Yin, Huan and Xu, Xuecheng and Lu, Sha and Chen, Xieyuanli and Xiong, Rong and Shen, Shaojie and Stachniss, Cyrill and Wang, Yue},
	journal={International Journal of Computer Vision},
	volume={132},
	number={8},
	pages={3139--3171},
	year={2024},
	publisher={Springer}
}

@inproceedings{bijelic2020seeing,
	title={Seeing through fog without seeing fog: Deep multimodal sensor fusion in unseen adverse weather},
	author={Bijelic, Mario and Gruber, Tobias and Mannan, Fahim and Kraus, Florian and Ritter, Werner and Dietmayer, Klaus and Heide, Felix},
	booktitle={Proceedings of the IEEE/CVF Conference on Computer Vision and Pattern Recognition},
	pages={11682--11692},
	year={2020}
}

@inproceedings{nakashima2024lidar,
	title={Lidar data synthesis with denoising diffusion probabilistic models},
	author={Nakashima, Kazuto and Kurazume, Ryo},
	booktitle={2024 IEEE International Conference on Robotics and Automation (ICRA)},
	pages={14724--14731},
	year={2024},
	organization={IEEE}
}

@inproceedings{wu2024text2lidar,
	title={Text2lidar: Text-guided lidar point cloud generation via equirectangular transformer},
	author={Wu, Yang and Zhang, Kaihua and Qian, Jianjun and Xie, Jin and Yang, Jian},
	booktitle={European Conference on Computer Vision},
	pages={291--310},
	year={2024},
	organization={Springer}
}

@article{ramesh2022hierarchical,
	title={Hierarchical text-conditional image generation with clip latents},
	author={Ramesh, Aditya and Dhariwal, Prafulla and Nichol, Alex and Chu, Casey and Chen, Mark},
	journal={arXiv preprint arXiv:2204.06125},
	volume={1},
	number={2},
	pages={3},
	year={2022}
}

@article{saharia2022photorealistic,
	title={Photorealistic text-to-image diffusion models with deep language understanding},
	author={Saharia, Chitwan and Chan, William and Saxena, Saurabh and Li, Lala and Whang, Jay and Denton, Emily L and Ghasemipour, Kamyar and Gontijo Lopes, Raphael and Karagol Ayan, Burcu and Salimans, Tim and others},
	journal={Advances in neural information processing systems},
	volume={35},
	pages={36479--36494},
	year={2022}
}

@inproceedings{rombach2022high,
	title={High-resolution image synthesis with latent diffusion models},
	author={Rombach, Robin and Blattmann, Andreas and Lorenz, Dominik and Esser, Patrick and Ommer, Bj{\"o}rn},
	booktitle={Proceedings of the IEEE/CVF conference on computer vision and pattern recognition},
	pages={10684--10695},
	year={2022}
}

@article{schuhmann2022laion,
	title={Laion-5b: An open large-scale dataset for training next generation image-text models},
	author={Schuhmann, Christoph and Beaumont, Romain and Vencu, Richard and Gordon, Cade and Wightman, Ross and Cherti, Mehdi and Coombes, Theo and Katta, Aarush and Mullis, Clayton and Wortsman, Mitchell and others},
	journal={Advances in neural information processing systems},
	volume={35},
	pages={25278--25294},
	year={2022}
}

@misc{coyo700m2023,
	title        = {COYO-700M: Large-scale Image-Text Pairs Dataset},
	author       = {Kakao Brain},
	year         = {2023},
	howpublished = {\url{https://github.com/kakaobrain/coyo-dataset}},
	note         = {Accessed: 2025-10-22}
}

@inproceedings{radford2021learning,
	title={Learning transferable visual models from natural language supervision},
	author={Radford, Alec and Kim, Jong Wook and Hallacy, Chris and Ramesh, Aditya and Goh, Gabriel and Agarwal, Sandhini and Sastry, Girish and Askell, Amanda and Mishkin, Pamela and Clark, Jack and others},
	booktitle={International conference on machine learning},
	pages={8748--8763},
	year={2021},
	organization={PmLR}
}

@inproceedings{liu2022design,
	title={Design guidelines for prompt engineering text-to-image generative models},
	author={Liu, Vivian and Chilton, Lydia B},
	booktitle={Proceedings of the 2022 CHI conference on human factors in computing systems},
	pages={1--23},
	year={2022}
}

@article{oppenlaender2025prompting,
	title={Prompting AI art: An investigation into the creative skill of prompt engineering},
	author={Oppenlaender, Jonas and Linder, Rhema and Silvennoinen, Johanna},
	journal={International journal of human--computer interaction},
	volume={41},
	number={16},
	pages={10207--10229},
	year={2025},
	publisher={Taylor \& Francis}
}

@inproceedings{caesar2020nuscenes,
	title={nuscenes: A multimodal dataset for autonomous driving},
	author={Caesar, Holger and Bankiti, Varun and Lang, Alex H and Vora, Sourabh and Liong, Venice Erin and Xu, Qiang and Krishnan, Anush and Pan, Yu and Baldan, Giancarlo and Beijbom, Oscar},
	booktitle={Proceedings of the IEEE/CVF conference on computer vision and pattern recognition},
	pages={11621--11631},
	year={2020}
}

@article{li2023self,
	title={Self-conditioned image generation via generating representations},
	author={Li, Tianhong and Katabi, Dina and He, Kaiming},
	journal={CoRR},
	year={2023}
}

@article{yu2024representation,
	title={Representation alignment for generation: Training diffusion transformers is easier than you think},
	author={Yu, Sihyun and Kwak, Sangkyung and Jang, Huiwon and Jeong, Jongheon and Huang, Jonathan and Shin, Jinwoo and Xie, Saining},
	journal={arXiv preprint arXiv:2410.06940},
	year={2024}
}

@inproceedings{zyrianov2022learning,
	title={Learning to generate realistic lidar point clouds},
	author={Zyrianov, Vlas and Zhu, Xiyue and Wang, Shenlong},
	booktitle={European Conference on Computer Vision},
	pages={17--35},
	year={2022},
	organization={Springer}
}

@inproceedings{caccia2019deep,
	title={Deep generative modeling of lidar data},
	author={Caccia, Lucas and Van Hoof, Herke and Courville, Aaron and Pineau, Joelle},
	booktitle={2019 IEEE/RSJ International Conference on Intelligent Robots and Systems (IROS)},
	pages={5034--5040},
	year={2019},
	organization={IEEE}
}

@inproceedings{hahner2021fog,
	title={Fog simulation on real LiDAR point clouds for 3D object detection in adverse weather},
	author={Hahner, Martin and Sakaridis, Christos and Dai, Dengxin and Van Gool, Luc},
	booktitle={Proceedings of the IEEE/CVF international conference on computer vision},
	pages={15283--15292},
	year={2021}
}

@inproceedings{teufel2022simulating,
	title={Simulating realistic rain, snow, and fog variations for comprehensive performance characterization of lidar perception},
	author={Teufel, Sven and Volk, Georg and Von Bernuth, Alexander and Bringmann, Oliver},
	booktitle={2022 IEEE 95th Vehicular Technology Conference:(VTC2022-Spring)},
	pages={1--7},
	year={2022},
	organization={IEEE}
}

@inproceedings{yang2024realistic,
	title={Realistic rainy weather simulation for lidars in carla simulator},
	author={Yang, Donglin and Cai, Xinyu and Liu, Zhenfeng and Jiang, Wentao and Zhang, Bo and Yan, Guohang and Gao, Xing and Liu, Si and Shi, Botian},
	booktitle={2024 IEEE/RSJ International Conference on Intelligent Robots and Systems (IROS)},
	pages={951--957},
	year={2024},
	organization={IEEE}
}

@article{kingma2013auto,
	title={Auto-encoding variational bayes},
	author={Kingma, Diederik P and Welling, Max},
	journal={arXiv preprint arXiv:1312.6114},
	year={2013}
}

@article{goodfellow2014generative,
	title={Generative adversarial nets},
	author={Goodfellow, Ian J and Pouget-Abadie, Jean and Mirza, Mehdi and Xu, Bing and Warde-Farley, David and Ozair, Sherjil and Courville, Aaron and Bengio, Yoshua},
	journal={Advances in neural information processing systems},
	volume={27},
	year={2014}
}

@article{ho2020denoising,
	title={Denoising diffusion probabilistic models},
	author={Ho, Jonathan and Jain, Ajay and Abbeel, Pieter},
	journal={Advances in neural information processing systems},
	volume={33},
	pages={6840--6851},
	year={2020}
}

@inproceedings{ran2024towards,
	title={Towards realistic scene generation with lidar diffusion models},
	author={Ran, Haoxi and Guizilini, Vitor and Wang, Yue},
	booktitle={Proceedings of the IEEE/CVF Conference on Computer Vision and Pattern Recognition},
	pages={14738--14748},
	year={2024}
}

@article{nichol2022point,
	title={Point-e: A system for generating 3d point clouds from complex prompts},
	author={Nichol, Alex and Jun, Heewoo and Dhariwal, Prafulla and Mishkin, Pamela and Chen, Mark},
	journal={arXiv preprint arXiv:2212.08751},
	year={2022}
}

@article{poole2022dreamfusion,
	title={Dreamfusion: Text-to-3d using 2d diffusion},
	author={Poole, Ben and Jain, Ajay and Barron, Jonathan T and Mildenhall, Ben},
	journal={arXiv preprint arXiv:2209.14988},
	year={2022}
}

@inproceedings{lin2023magic3d,
	title={Magic3d: High-resolution text-to-3d content creation},
	author={Lin, Chen-Hsuan and Gao, Jun and Tang, Luming and Takikawa, Towaki and Zeng, Xiaohui and Huang, Xun and Kreis, Karsten and Fidler, Sanja and Liu, Ming-Yu and Lin, Tsung-Yi},
	booktitle={Proceedings of the IEEE/CVF conference on computer vision and pattern recognition},
	pages={300--309},
	year={2023}
}

@inproceedings{luo2021diffusion,
	title={Diffusion probabilistic models for 3d point cloud generation},
	author={Luo, Shitong and Hu, Wei},
	booktitle={Proceedings of the IEEE/CVF conference on computer vision and pattern recognition},
	pages={2837--2845},
	year={2021}
}

@inproceedings{wu2023sketch,
	title={Sketch and text guided diffusion model for colored point cloud generation},
	author={Wu, Zijie and Wang, Yaonan and Feng, Mingtao and Xie, He and Mian, Ajmal},
	booktitle={Proceedings of the IEEE/CVF International Conference on Computer Vision},
	pages={8929--8939},
	year={2023}
}

@inproceedings{liu2025fshnet,
	title={FSHNet: Fully Sparse Hybrid Network for 3D Object Detection},
	author={Liu, Shuai and Cui, Mingyue and Li, Boyang and Liang, Quanmin and Hong, Tinghe and Huang, Kai and Shan, Yunxiao},
	booktitle={Proceedings of the Computer Vision and Pattern Recognition Conference},
	pages={8900--8909},
	year={2025}
}

@article{zhou2024layout,
	title={Layout-your-3d: Controllable and precise 3d generation with 2d blueprint},
	author={Zhou, Junwei and Li, Xueting and Qi, Lu and Yang, Ming-Hsuan},
	journal={arXiv preprint arXiv:2410.15391},
	year={2024}
}

@article{zhou2025layoutdreamer,
	title={Layoutdreamer: Physics-guided layout for text-to-3d compositional scene generation},
	author={Zhou, Yang and He, Zongjin and Li, Qixuan and Wang, Chao},
	journal={arXiv preprint arXiv:2502.01949},
	year={2025}
}

@inproceedings{liu2022compositional,
	title={Compositional visual generation with composable diffusion models},
	author={Liu, Nan and Li, Shuang and Du, Yilun and Torralba, Antonio and Tenenbaum, Joshua B},
	booktitle={European conference on computer vision},
	pages={423--439},
	year={2022},
	organization={Springer}
}

@article{podell2023sdxl,
	title={Sdxl: Improving latent diffusion models for high-resolution image synthesis},
	author={Podell, Dustin and English, Zion and Lacey, Kyle and Blattmann, Andreas and Dockhorn, Tim and M{\"u}ller, Jonas and Penna, Joe and Rombach, Robin},
	journal={arXiv preprint arXiv:2307.01952},
	year={2023}
}

@inproceedings{milioto2019rangenet++,
	title={Rangenet++: Fast and accurate lidar semantic segmentation},
	author={Milioto, Andres and Vizzo, Ignacio and Behley, Jens and Stachniss, Cyrill},
	booktitle={2019 IEEE/RSJ international conference on intelligent robots and systems (IROS)},
	pages={4213--4220},
	year={2019},
	organization={IEEE}
}

@inproceedings{qu2024conditional,
	title={A conditional denoising diffusion probabilistic model for point cloud upsampling},
	author={Qu, Wentao and Shao, Yuantian and Meng, Lingwu and Huang, Xiaoshui and Xiao, Liang},
	booktitle={Proceedings of the IEEE/CVF Conference on Computer Vision and Pattern Recognition},
	pages={20786--20795},
	year={2024}
}

@article{ho2022classifier,
	title={Classifier-free diffusion guidance},
	author={Ho, Jonathan and Salimans, Tim},
	journal={arXiv preprint arXiv:2207.12598},
	year={2022}
}

@article{wang2023patch,
	title={Patch diffusion: Faster and more data-efficient training of diffusion models},
	author={Wang, Zhendong and Jiang, Yifan and Zheng, Huangjie and Wang, Peihao and He, Pengcheng and Wang, Zhangyang and Chen, Weizhu and Zhou, Mingyuan and others},
	journal={Advances in neural information processing systems},
	volume={36},
	pages={72137--72154},
	year={2023}
}

@article{zhu2025domainstudio,
	title={Domainstudio: Fine-tuning diffusion models for domain-driven image generation using limited data},
	author={Zhu, Jingyuan and Ma, Huimin and Chen, Jiansheng and Yuan, Jian},
	journal={International Journal of Computer Vision},
	volume={133},
	number={10},
	pages={7012--7036},
	year={2025},
	publisher={Springer}
}

@article{oquab2023dinov2,
	title={Dinov2: Learning robust visual features without supervision},
	author={Oquab, Maxime and Darcet, Timoth{\'e}e and Moutakanni, Th{\'e}o and Vo, Huy and Szafraniec, Marc and Khalidov, Vasil and Fernandez, Pierre and Haziza, Daniel and Massa, Francisco and El-Nouby, Alaaeldin and others},
	journal={arXiv preprint arXiv:2304.07193},
	year={2023}
}

@article{liao2022kitti,
	title={Kitti-360: A novel dataset and benchmarks for urban scene understanding in 2d and 3d},
	author={Liao, Yiyi and Xie, Jun and Geiger, Andreas},
	journal={IEEE Transactions on Pattern Analysis and Machine Intelligence},
	volume={45},
	number={3},
	pages={3292--3310},
	year={2022},
	publisher={IEEE}
}

@article{sauer2021projected,
	title={Projected gans converge faster},
	author={Sauer, Axel and Chitta, Kashyap and M{\"u}ller, Jens and Geiger, Andreas},
	journal={Advances in Neural Information Processing Systems},
	volume={34},
	pages={17480--17492},
	year={2021}
}

@inproceedings{he2023grad,
	title={Grad-pu: Arbitrary-scale point cloud upsampling via gradient descent with learned distance functions},
	author={He, Yun and Tang, Danhang and Zhang, Yinda and Xue, Xiangyang and Fu, Yanwei},
	booktitle={Proceedings of the IEEE/CVF Conference on Computer Vision and Pattern Recognition},
	pages={5354--5363},
	year={2023}
}

@inproceedings{zhang2023adding,
	title={Adding conditional control to text-to-image diffusion models},
	author={Zhang, Lvmin and Rao, Anyi and Agrawala, Maneesh},
	booktitle={Proceedings of the IEEE/CVF international conference on computer vision},
	pages={3836--3847},
	year={2023}
}

@inproceedings{li2019pu,
	title={Pu-gan: a point cloud upsampling adversarial network},
	author={Li, Ruihui and Li, Xianzhi and Fu, Chi-Wing and Cohen-Or, Daniel and Heng, Pheng-Ann},
	booktitle={Proceedings of the IEEE/CVF international conference on computer vision},
	pages={7203--7212},
	year={2019}
}

@inproceedings{behley2019semantickitti,
	title={Semantickitti: A dataset for semantic scene understanding of lidar sequences},
	author={Behley, Jens and Garbade, Martin and Milioto, Andres and Quenzel, Jan and Behnke, Sven and Stachniss, Cyrill and Gall, Jurgen},
	booktitle={Proceedings of the IEEE/CVF international conference on computer vision},
	pages={9297--9307},
	year={2019}
}

@inproceedings{manivasagam2020lidarsim,
	title={Lidarsim: Realistic lidar simulation by leveraging the real world},
	author={Manivasagam, Sivabalan and Wang, Shenlong and Wong, Kelvin and Zeng, Wenyuan and Sazanovich, Mikita and Tan, Shuhan and Yang, Bin and Ma, Wei-Chiu and Urtasun, Raquel},
	booktitle={Proceedings of the IEEE/CVF Conference on Computer Vision and Pattern Recognition},
	pages={11167--11176},
	year={2020}
}

@article{leng2025repae,
	title={REPA-E: Unlocking VAE for End-to-End Tuning with Latent Diffusion Transformers},
	author={Xingjian Leng and Jaskirat Singh and Yunzhong Hou and Zhenchang Xing and Saining Xie and Liang Zheng},
	year={2025},
	journal={arXiv preprint arXiv:2504.10483},
}

@article{salimans2022progressive,
	title={Progressive distillation for fast sampling of diffusion models},
	author={Salimans, Tim and Ho, Jonathan},
	journal={arXiv preprint arXiv:2202.00512},
	year={2022}
}

@inproceedings{qu2025end,
	title={An end-to-end robust point cloud semantic segmentation network with single-step conditional diffusion models},
	author={Qu, Wentao and Wang, Jing and Gong, YongShun and Huang, Xiaoshui and Xiao, Liang},
	booktitle={Proceedings of the Computer Vision and Pattern Recognition Conference},
	pages={27325--27335},
	year={2025}
}

@article{qu2025robust,
	title={Robust Single-Stage Fully Sparse 3D Object Detection via Detachable Latent Diffusion},
	author={Qu, Wentao and Mei, Guofeng and Wang, Jing and Wu, Yujiao and Huang, Xiaoshui and Xiao, Liang},
	journal={arXiv preprint arXiv:2508.03252},
	year={2025}
}

@inproceedings{stearns2024curvecloudnet,
	title={Curvecloudnet: Processing point clouds with 1d structure},
	author={Stearns, Colton and Fu, Alex and Liu, Jiateng and Park, Jeong Joon and Rempe, Davis and Paschalidou, Despoina and Guibas, Leonidas J},
	booktitle={Proceedings of the IEEE/CVF Conference on Computer Vision and Pattern Recognition},
	pages={27981--27991},
	year={2024}
}
}

\end{document}